\documentclass{article} 
\usepackage{arxiv,times}

\usepackage{amsmath,amsfonts,bm}

\def\eqref#1{equation~\ref{#1}}

\def\1{\bm{1}}

\DeclareMathAlphabet{\mathsfit}{\encodingdefault}{\sfdefault}{m}{sl}
\SetMathAlphabet{\mathsfit}{bold}{\encodingdefault}{\sfdefault}{bx}{n}

\usepackage{hyperref}
\usepackage{url}
\usepackage{enumitem}
\usepackage{booktabs}  
\usepackage{tabularx}   
\usepackage{array}     
\usepackage{graphicx}
\usepackage{float}
\usepackage{makecell}
\usepackage{natbib}
\usepackage{placeins}

\title{Draw it like Euclid:  Teaching transformer models to generate CAD profiles using ruler and compass construction steps}

\author{Siyi Li, Joseph G. Lambourne,  Longfei Zhang,  Pradeep Kumar Jayaraman,  Karl. D.D. Willis \\
Autodesk Research\\
}

\def\etal{\emph{et al.~}}

\begin{document}

\maketitle

\begin{abstract}
We introduce a new method of generating Computer Aided Design (CAD) profiles via a sequence of simple geometric constructions including curve offsetting, rotations and intersections.  These sequences start with geometry provided by a designer and build up the points and curves of the final profile step by step. We demonstrate that adding construction steps between the designer's input geometry and the final profile improves generation quality in a similar way to the introduction of a chain of thought in language models.  Similar to the constraints in a parametric CAD model, the construction sequences reduce the degrees of freedom in the modeled shape to a small set of parameter values which can be adjusted by the designer, allowing parametric editing with the constructed geometry evaluated to floating point precision.  In addition we show that applying reinforcement learning to the construction sequences gives further improvements over a wide range of metrics, including some which were not explicitly optimized.
\end{abstract}

\begin{figure}[H]
    \centering
    \includegraphics[width=\textwidth]{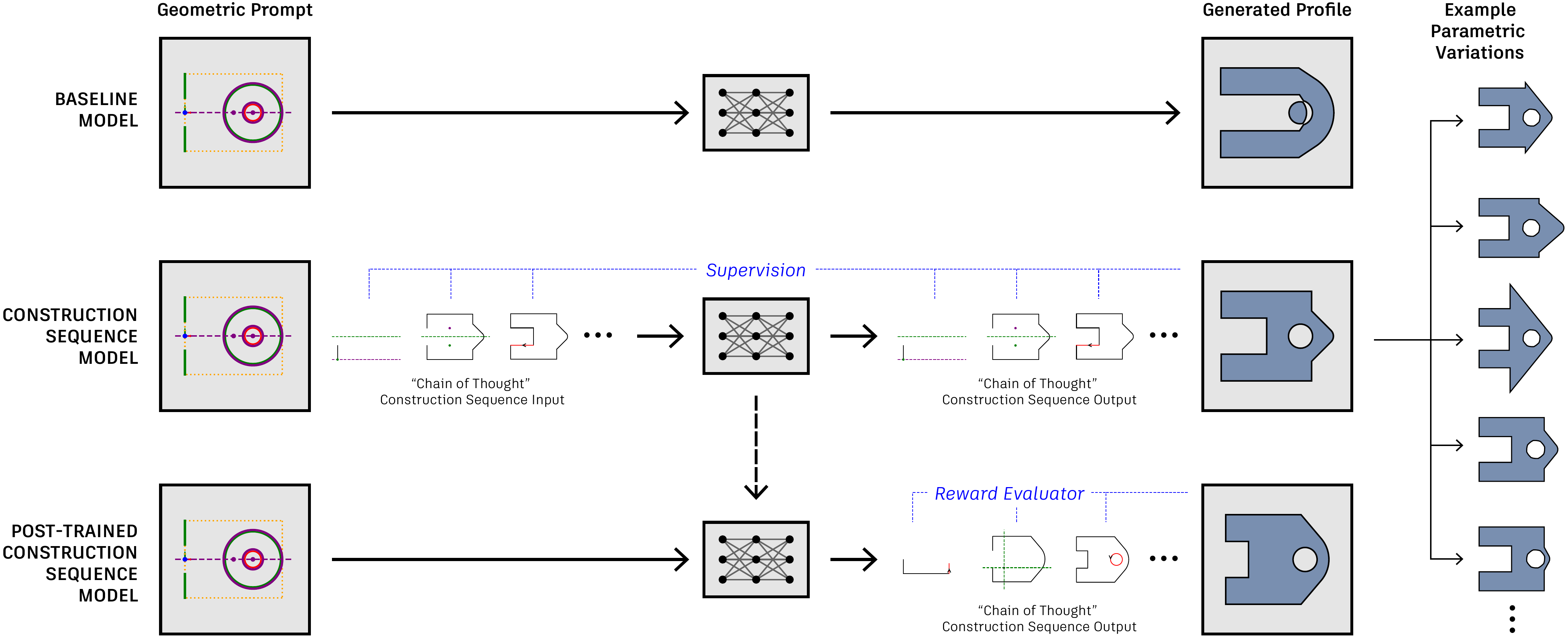}
    \caption{Our model (middle) generates CAD profiles through a sequence of ruler, compass, and protractor constructions, starting from the designer’s initial ``prompt" geometry (left) and building step by step toward the final profile. Our approach generates profiles that more accurately match the designer’s input and contain fewer self-intersections, compared to a baseline model (top) which omits construction steps and maps directly from geometric prompt to profile. We further refine the construction sequence model during post-training (bottom), where rewards help guide the generation of valid construction sequences. As the sequences encode parametric relationships, families of related profiles (right) can be created from a single construction trace.}
    \label{fig:construction_steps_teaser}
\end{figure}

\section{Introduction}
\label{introduction}

Computer Aided Design (CAD) tools play a key role in shaping nearly all manufactured objects. CAD models are created by specifying collections of lines, arcs and circles which enclose 2d regions called profiles.  These can then be extruded or revolved to define solid volumes, which can be combined using boolean operations to build complex shapes.  Geometric constraints can be added to the profiles, enforcing relationships between the curves.  For example, lines can be constrained to be parallel, circles concentric and curves can be forced to meet tangentially.   The distances and angles between specific curves can also be controlled using parameters, which can be adjusted to modify the shape while maintaining critical aspects of the design like symmetries, regular patterns and constant thicknesses.

While machine learning models for the generation of 2d CAD geometry have shown great advances in recent years, these methods have two main limitations.  Firstly, all current methods suffer from limited accuracy.   Methods which create geometry as a sequence of discrete tokens representing quantized points or coordinates \cite{Ari2020sketchgraphs, willis2021, Ganin2021, Para2021sketchgen, Wu2023iconshop} have their precision limited by the spacing of the quantization grid, while diffusion methods \cite{xu2024brepgen, fan2024neuronurbs, lee2025brepdiff} have accuracy limited by the convergence of the diffusion process and the decoding of latent vectors.  Secondly, while some methods can predict constraints and dimensions, this is done as a second step after the geometry has been generated.   An external constraint solver is then required to post-process the curve geometry and produce the final shape.  It has been shown that applying generated constraints in post-process can frequently move the geometry \cite{Para2021sketchgen,Ganin2021} which is undesirable when the changes significantly alter the design \cite{casey2025aligningconstraintgenerationdesign}.

In this work we investigate an alternative strategy for generating 2d profiles, using a unified sequence which defines both the geometry and shape properties which would usually be enforced by constraints.  Inspired by recent work in language models, which show that an intermediate chain of thought (CoT) can greatly improve the accuracy of the final output \cite{Wei2022, Kojima2022}, we wonder whether the generation of some intermediate geometry might assist with CAD generations tasks.   We notice that when CAD designers build up shapes, they often start by sketching some intermediate ``construction geometry" which defines important aspects of a design like symmetry lines or construction circles.  Constraints are then added between the construction geometry and profile curves to enforce properties like coincidence and symmetries.   Inside the geometric constraint solver, the graph formed by the geometry and constraints is recursively processed to yield a fine grained sequence of simple geometric constructions which builds up the final shape \cite{Owen1991,Bouma1995}.   These construction sequences play a role similar to the algorithm execution traces which have been employed to great effect in a variety of search problems \cite{Yang2022ChainOT,Lehnert2024BeyondAB, Gandhi2024StreamOS}.     

In this paper we conduct experiments training transformer models on intermediate construction sequences of ruler, compass and protractor construction steps similar to those used in geometric constraint solvers. As the high level task of shape generation is broken down into small atomic steps with closed form solutions, the performance of the network at solving these subtasks and combining them to build a consistent ``CAD program" can be measured separately.  The shapes can be controlled using  ``prompt geometry" provided at the start of the sequence and used as inputs to subsequent constructions. Once generated, the sequences can be replayed with floating point precision, allowing accurate values known at inference time to be propagated through the constructions to build the final profile.  The generated geometry is parameterized using a small number values, which can be varied when the sequences are replayed allowing parametric edits to the shape as shown in the right of  Figure \ref{fig:construction_steps_teaser}.

We demonstrate quantitatively that the introduction of construction sequences improves the performance of the generative models, reducing self-intersections and proving enhanced adherence to the design requirements. In addition, we show that applying reinforcement learning over the entire construction sequence can further improve results as shown in the language modeling case \cite{shao2024grpo, GouDeepSeekR1}.

Our contributions can be summarized as follows
\begin{itemize}
    \item We introduce a domain specific language which builds 2d profile geometry as a sequence of ruler, compass and protractor constructions steps.   The construction steps can be replayed with floating point accuracy, allowing parametric editing of the shape.
    \item We show that geometry generated with sequences which include these intermediate construction steps have fewer self-intersections, superior accuracy when auto-completing partial designs and better adherence to other design requirements like symmetry lines.
    \item We show that reinforcement learning, with reward functions which discourage self-intersecting geometry, leads to improvement over a wide range of metrics, many of which are not explicitly optimized.
\end{itemize}

\section{Related work}
\label{related_work}

\subsection{Algorithm traces}
A number of works have explored training neural networks to mimic the logical steps conducted by heuristic algorithms. Vinyals \etal \cite{vinyals2015pointer} showed that RNNs could replicate some basic geometric algorithms like finding the convex hull and building a delaunay triangulation.  Yang \etal \cite{Yang2022ChainOT} experimented with monte carlo tree search traces for maze navigation, robotic manipulation and Atari games.   Lehnert \etal \cite{Lehnert2024BeyondAB} used traces from $A^{*}$ search to learn to solve mazes and Gandhi \etal \cite{Gandhi2024StreamOS} showed how search and backtracking capabilities could be used to play the game  Countdown.  In this work we investigate the applicability of these techniques to CAD,  utilizing algorithm traces similar to those used in geometric constraint solvers.

\subsection{Sketch generation}

The availability of large scale constrained sketch datasets \cite{Ari2020sketchgraphs,Ganin2021} opened the task of CAD sketch generation and sketch auto-constraining to the community. Seff \etal \cite{Ari2020sketchgraphs} presented an autoregressive model based on message passing networks which generated parametric sketches by iteratively predicted the edges and node attributes of the constraint graph.  An external geometric constraint solver was then required to construct the final geometry and the complexity of the resulting sketches was limited. Willis \etal \cite{willis2021}  showed that unconditional generation of 2d sketches was possible by first generating a list of points and then using PointerNetworks \cite{vinyals2015pointer} to group these to define curves.   Several prior works introduce models that first generate 2d curves and then, in a second step, used PointerNetworks to predict constraints and dimensions conditioned on this geometry \cite{Ganin2021,Para2021sketchgen,Seff2021VitruvionAG}.  Because these architectures predict design intent only after the geometry has been generated, they cannot leverage the constraints to guide the curve placement. Instead, an external constraint solver must be applied as a post-process. This can shift the positions of sketch geometry \cite{Ganin2021,Para2021sketchgen,casey2025aligningconstraintgenerationdesign}, revealing that the initial curve placement did not reflect the intended design. In contrast, the construction sequence representation introduced here enables design intent to be predicted before geometry is constructed, allowing the network to incorporate it directly when placing points and curves.

\subsection{Chain of thought in CAD}

Text2CAD \cite{khan2024textcad} used parametric recipes from the DeepCAD dataset \cite{wu2021deepcad} to define a natural language description of the CAD modeling features used to created the solids.  
CAD-Coder \cite{guan2025cadcoder} used Deepseek-V3 to convert these descriptions into a natural language CoT followed by CadQuery code. 
A Qwen2.5-7B-Instruct model was fine tuned on this data and the GRPO reinforcement learning algorithm with a chamfer distance based reward function was used to further enhance results.  Seek-CAD \cite{li2025seekcad} used DeepSeek-R1 to generate a natural language CoT and CAD commands from a text description.  The CoT was passed to Gemini-2.0 along with images of the generated CAD model to provide visual feedback in an iterative refinement loop.  The natural language CoTs employed by these models were used as an auxiliary representation along side the executable code.  While they includes statements related to design intent, these are not defined in a formal language which can be directly executed by CAD kernels.

\subsection{Reinforcement Learning for CAD}
A number of recent papers have shown promising results applying reinforcement learning to a variety of CAD related tasks. 
Casey \etal\cite{casey2025aligningconstraintgenerationdesign} fine tuned a CAD the auto-constrainer model from \cite{Seff2021VitruvionAG} with a number of RL algorithms.  This was shown to improve the fraction of entities fully defined by the constraints while reducing the fraction of sketches where geometry moved when the constraints were applied.  RL-CAD \cite{yin2025rlcad} studied the recovery of a parametric feature recipe from B-Rep models. An Actor-Critic network selected faces of the target B-Rep to extrude or revolve and rewards were based on the similarity between the recovered and target shapes. 
CADCrafter \cite{chen2025cadcrafter} used Direct Preference Optimization (DPO) to improve the performance of an image to CAD command sequence network and      
Cadrille  \cite{kolodiazhnyi2025cadrille} studied the use of reinforcement learning to improve CAD reconstruction from point clouds. 

\section{Data}
\label{data}

\subsection{Dataset creation} \label{subsection:dataset_creation}

While large scale datasets of CAD sketch geometry exist \cite{Ari2020sketchgraphs}, the curves in these datasets typically do not form closed loops as required for downstream modeling operations such as constructing extrusions or solids of revolution.  Our training data comprises closed profiles which are derived from the CAD models in the ABC dataset \cite{koch2019abc}.  The Open Cascade \cite{opencascade} modeling kernel is used to create the profiles, by slicing each B-Rep solid with 5 equally spaced section planes with normals along each of the three  coordinate system axes.  Disjoint regions are separated so that each extracted profile has one outer loop and zero or more inner loops. 
This results in closed profile loops consisting of line segments, arcs and circles.  The data deduplicated procedure converted each profile into a graph with nodes as the vertices and curves as edges.   The edges were labeled with the curve type and nodes were labeled using the vertex coordinates, quantized into 8x8 bins. The Weisfeiler Lehman graph hash \cite{Shervashidze2011} was then computed and profiles with duplicated hashes were removed.  The data was then split into $95\%$ train, $3\%$ validation and $2\%$ test.  

The extraction of the construction sequences from the raw profile geometry utilizes a set of simple heuristic algorithms which are described briefly here. A detailed description of each of the algorithm's phases can then be found in Appendix \ref{appendix:profile2prompt_algorithm}.  To simulate the input of a designer, we start our sequences with information used for shape control which we refer to as a ``geometric prompt".  Rather than auto-completing a sketch from a random subset of sketch geometry as in  \cite{Seff2021VitruvionAG}, we extract line segments from the convex hull and the positions of internal circular loops.  The area and bounding box of the profile, along with any symmetry lines are also included. Next in an ``analysis phase", we identify geometric relationships between curves such as parallel lines, concentric circles and fillet arcs.   These relationships are translated into construction steps, such as curve offsetting and filleting operations. We then built a bipartite dataflow graph in which nodes represent geometry and construction steps and directed edges represent how geometry flows into and out of the operations.  Initially this graph will contain cycles and redundant branches, which are removed in a ``graph simplification phase".  The construction sequence is then obtained using a lexicographical topological sort, in which the order of the curves in the final profile is used to resolve ambiguities in the topological sort order.  The data extraction process yielded a total of 318,208 profiles with corresponding construction sequences.


\subsection{Learned sequences} \label{subsection:learned sequences}
Our experiments utilize sequences with the following three components. The sequences start with the ``geometric prompt" which is used to control the shape.  Next we include the construction steps, which act like a chain of thought, starting with the prompt geometry and constructing the points and curves required to define the final profile. Finally we have the profile geometry, which is analogous to the final answer returned by a reasoning LLM.  

The construction steps represent simple geometric operations such as curve offsetting, curve-curve intersections, curve reversal, mirroring points over symmetry lines and the construction of fillet arcs.   A few examples of supported construction steps are provided in Table~\ref{tab:construction_steps_examples}, and the remainder are listed in Table~\ref{tab:construction_steps_contd} of Appendix \ref{appendix:construction_step_geometry}.   Each construction step has an operation type, a list of input geometry and a list of output geometry.  The output geometry of one step can be utilized as the input to subsequent steps, building up a description of the dataflow graph.    The construction steps are ordered such that the curves are created in the order they appear in the profile, with the first curve chosen so that its end point is closest to the bottom left hand corner.  Details of the  domain specific language (DSL) and tokenization used to encode the sequences are in Appendix \ref{appendix:tokenization}.

\begin{table}[!htb]
    \centering
    \caption{Ruler and compass construction steps. Examples of input geometries to the construction steps are shown in blue and their output in red.}
    \label{tab:construction_steps_examples}
    \renewcommand{\arraystretch}{0.5}
    \begin{tabular}{@{}p{5cm} p{5cm} p{3cm}@{}}
        \toprule
        \textbf{Description} & \textbf{Explanation} & \textbf{Example} \\
        \midrule
        \begin{tabular}[t]{@{}l@{}}
        \texttt{CircleOffsetCircle} \\
        \textbf{Input:} \texttt{circle\textsubscript{1}, offset} \\
        \textbf{Output:} \texttt{circle\textsubscript{2}}
        \end{tabular}
        &
        Given an oriented circle and a positive offset distance, find and return the offset circle.
        &
        \raisebox{-0.75\height}{\includegraphics[width=2cm]{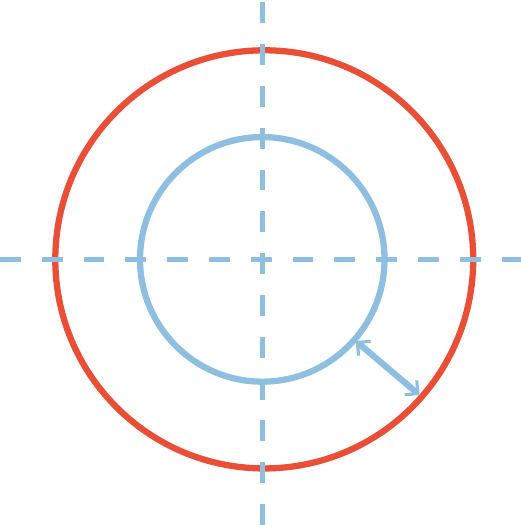}}
        \\

        \midrule
        \begin{tabular}[t]{@{}l@{}}
        \texttt{LineXLine} \\
        \textbf{Input:} \texttt{line\textsubscript{1}, line\textsubscript{2}} \\
        \textbf{Output:} \texttt{point}
        \end{tabular}
        &
        Given two lines, find and return their intersection point.
        &
        \raisebox{-0.75\height}{\includegraphics[width=2cm]{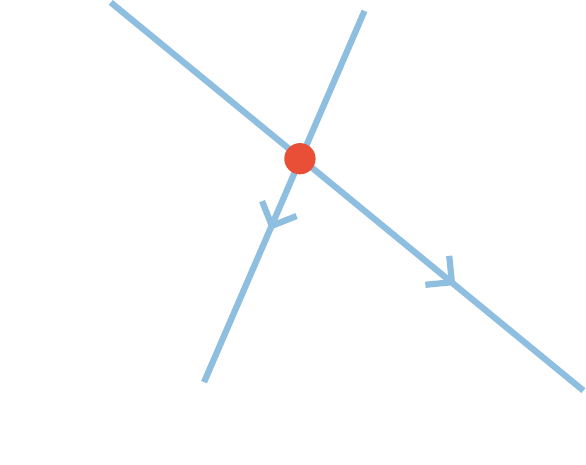}}
        \\
        \midrule
        \begin{tabular}[t]{@{}l@{}}
        \texttt{LineOffsetLine} \\
        \textbf{Input:} \texttt{line\textsubscript{1}, offset} \\
        \textbf{Output:} \texttt{line\textsubscript{2}}
        \end{tabular}
        &
        Given a directed line and an offset distance, find and return the line offset from this line to the left hand side by the offset distance.
        &
        \raisebox{-0.75\height}{\includegraphics[width=2cm]{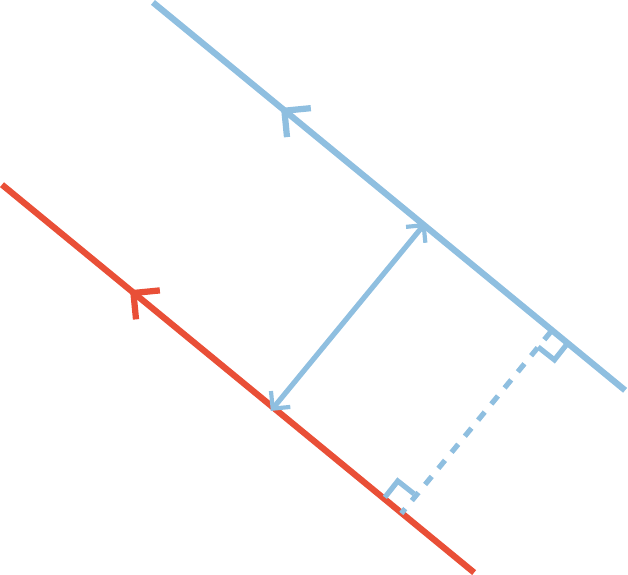}}
        \\

        \midrule
        \begin{tabular}[t]{@{}l@{}}
        \texttt{LineXCicle} \\
        \textbf{Input:} \texttt{line, circle} \\
        \textbf{Output:} \texttt{point(s)}
        \end{tabular}
        &
        Given a line and a circle, find and return the intersection point(s).
        &
        \raisebox{-0.75\height}{\includegraphics[width=2cm]{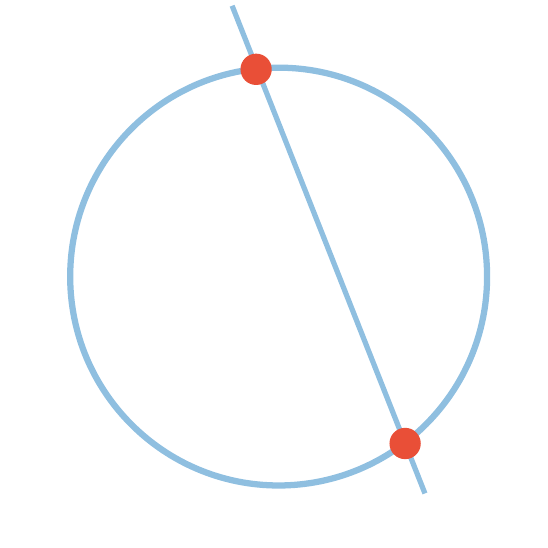}}
        \\
        \bottomrule
    \end{tabular}
\end{table}

\section{Method}
\label{method}

\subsection{Supervised learning}


We train an autoregressive decoder only transformer on the sequences described in \ref{subsection:learned sequences} with a cross entropy loss. In our experiments we use 8 heads, a depth of 8, and embedding dimension of 1024, and an attention head dimension of 128.  We use the Adam optimizer \cite{Kingma2014AdamAM} with a learning rate of $3e-4$ and dropout of 0.1.  Training was performed on 4 RTX 6000 GPUs. 

Two variants of the model were trained.  A baseline model which includes only the information in the geometric prompt and then the geometry of the final profile, and a construction sequence model which additionally includes the intermediate construction steps.   A comparison of the performance of these models is given in Section \ref{results:quantititive}

\subsection{Reinforcement learning (RL)}
\label{subsection:rl_method}
The fine-tuning of the profile generation model can be formulated as follows; the profile generation model $\pi_{\theta}(\tau|x)$ generates a profile sequence $\tau$ for geometric prompt $x$. Given a set of geometric prompts $D=\{x_i\}^N_i$ and reward function $r$ that provides a scalar value $r(x,\tau)$ which evaluates the quality of a profile sequence $\tau$ and how well it satisfies the geometric constraints prescribed by $x$. RL finetuning proceeds by maximizing the expected rewards 

\begin{equation}
    \mathop{\max}\limits_{\theta} \mathbb{E}_{x\sim\rho}\mathbb{E}_{\tau\sim\pi_{\theta}(\cdot|x)}[r(x,\tau)],
\end{equation}

where $\rho$ represents the distribution of the geometric prompts, and $\theta \in \mathbb{R}^{d}$ denotes for the training parameters of the profile generation model.


\subsubsection{Reward design}
\label{subsection:reward_design}
While reinforcement learning from human feedback (RLHF) tends to involve vague and subjective human preferences, the task of CAD profile generation allows for direct and objective evaluation of the generated profile sequences through measurable validity metrics, which eliminates the need for a learned reward model. The rewards used for RL are defined as follows:
\begin{itemize}[leftmargin=12pt]
    \item Reward for syntactically valid profile sequence $\tau$: 
    \begin{flushleft}
    $r_{\text{no self-intersection}}$: fraction of profiles without self-intersecting curves,
    
    $r_{\text{no short edges}}$: fraction of profiles without edges shorter than a predefined minimum length,
    \end{flushleft}
    \item Reward for syntactically invalid profile sequence:
    \begin{flushleft}
    $r_{\text{invalid profile}}$: penalty for generated sequences that produce syntactically invalid DSL code and cannot be detokenized.
    \end{flushleft}
\end{itemize}

\subsubsection{Policy gradient methods}

We focus on three sequence-level policy gradient methods: ReMax \citep{li2023remax}, GRPO \citep{shao2024grpo}, and RLOO \citep{ahmadian2024rloo}. These methods compute policy gradients using the total log-probability of the generated sequence and apply a REINFORCE-style estimator with learned baselines \citep{lex2001baseline}. In contrast to PPO \citep{schulman2017ppo}, which performs token-level policy updates using individual log-probabilities at each decoding step, these approaches operate at the sequence level and eliminates the need for a separately trained value network. We apply them to fine-tune the profile generation policy.

\noindent \textbf{ReMax} \citep{li2023remax} uses the reward of a sequence generated by greedily decoding from the policy network as a baseline to normalize the rewards of sequences sampled stochastically. 

With sequences $\tau$ sampled from policy $\pi_{\theta}(\tau|x)$, the ReMax baseline $b_{\theta, \text{ReMax}}$ is $\text{argmax}(\pi_{\theta}(\tau|x))$, and the policy gradient objective for ReMax is:

\begin{equation}
    \mathbb{E}_{\tau \sim \pi_{\theta}(\cdot \mid x)}
    \Big[
         \big( r(x, \tau) - b_{\theta, \text{ReMax}} \big) \cdot \nabla_{\theta} \log \pi_{\theta}(\tau \mid x)
    \Big]
\end{equation}

\noindent \textbf{Group Relative Policy Optimization (GRPO)} \citep{shao2024grpo} samples a group of $G$ individual profile sequences $\{\tau\}^G_{g=1}$ for every geometric prompt $x$. The advantage of the $g$-th profile sequence is calculated by normalizing the group-level rewards $\{r(x, \tau_g)\}^G_{g=1}$:
\begin{equation}
    A_g = \frac{r(x, \tau_g) - \text{mean}(\{r(x, \tau_g)\}^G_{g=1})}{\text{std}(\{r(x, \tau_g)\}^G_{g=1})}.
\end{equation}

The GRPO policy gradient objective is:

\begin{equation}
    \mathbb{E}_{\{\tau_g\}_{g=1}^G \sim \pi_{\theta}(\cdot \mid x)} 
    \Big[ 
        \text{min}(\psi_gA_g, \text{clip}\big(\psi_g, 1-\epsilon, 1+\epsilon\big)A_g) - \beta D_{\text{KL}}(\pi_{\theta}\|\pi_{\text{ref}})
    \Big],
\end{equation}

where:

\begin{equation}
  \begin{aligned}
    \psi_g
      &= \frac{\nabla_\theta \pi_\theta(\tau_g \mid x)}
              {\pi_{\text{ref}}(\tau_g \mid x)},
      \qquad
    D_{\mathrm{KL}}\!\bigl(\pi_\theta \,\|\, \pi_{\text{ref}}\bigr)
      = \frac{1}{\psi_g} + \text{log}\psi_g - 1
  \end{aligned}
\end{equation}
\noindent \textbf{REINFORCE-Leave-One-Out (RLOO)} \citep{ahmadian2024rloo} samples a group of $G$ individual profile sequences $\{\tau\}^G_{g=1}$ for every geometric prompt $x$. The reward for each sample within a group $r_g$ serves all other samples as a baseline, resulting in the policy gradient objective as follows:

\begin{equation}
    \mathbb{E}_{\{\tau_g\}_{g=1}^G \sim \pi_{\theta}(\cdot \mid x)} 
    \Big[ 
        \frac{1}{G} \sum_{g=1}^G 
        [
        \big( 
            r_g - \frac{1}{G - 1} \sum_{i \ne g} r_i 
        \big) 
        \cdot \nabla_{\theta} \log \pi_{\theta}(\tau_g \mid x)
        ] 
    \Big].
\end{equation}
\section{Results}

\subsection{Shape control via geometric prompting}
\label{appendix:shape_control_from_geometric_prompt}
Geometric prompts contain information which the designer can use to control the generated profile shape.  Here we show some qualitative examples of shape control.   In each of the figures in this section, a single value in the geometric prompt is varies and the other are held  fixed at the values shown in Table \ref{tab:geom_prompt_standard}. The positions of the line segments which the profile should match are shown in green, symmetry lines are marked with  dashed purple lines and the center of gravity shown as a purple dot.


\begin{table}[!ht]
    \centering
    \caption{Default properties used in the geometric prompt in this section.}
    \label{tab:geom_prompt_standard}
    \renewcommand{\arraystretch}{0.5}
    \begin{tabular}{@{}p{0.35\textwidth} p{0.65\textwidth}}
    \toprule
    \textbf{Property} & \textbf{Value}  \\
    \midrule
    
    Area & $0.4$ \\
    Symmetry & Vertical symmetry line \\
    Tangent continuous vertex fraction & $0.5$ \\
    Number of edges & $16$ \\
    Center of gravity & $0.0, 0.0$ \\
    Top width & $0.1$ \\
        
    \bottomrule
    \end{tabular}
\end{table}

\begin{figure}[ht]
  \centering
  \sbox0{\includegraphics{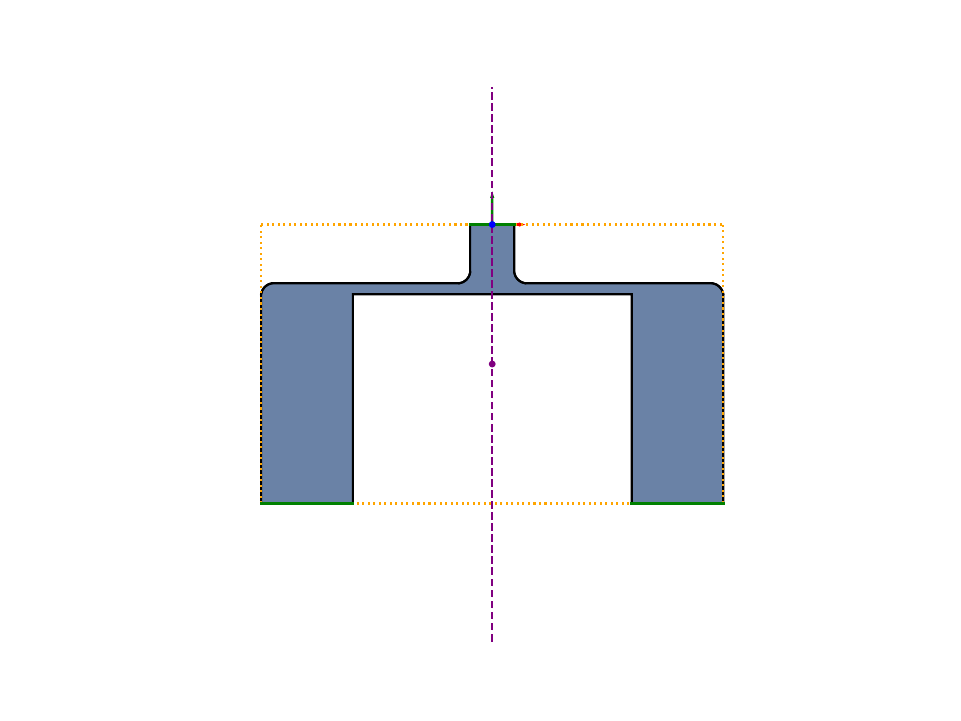}}
  \begin{tabular}{ccc}
  \includegraphics[width=0.32\textwidth,clip,trim={0.2\wd0 0.2\wd0 0.2\wd0 0.2\wd0}]{figures/geometric_prompt/Area/781_area_0.2_first_generation_00.pdf} &
  \includegraphics[width=0.32\textwidth,clip,trim={0.2\wd0 0.2\wd0 0.2\wd0 0.2\wd0}]{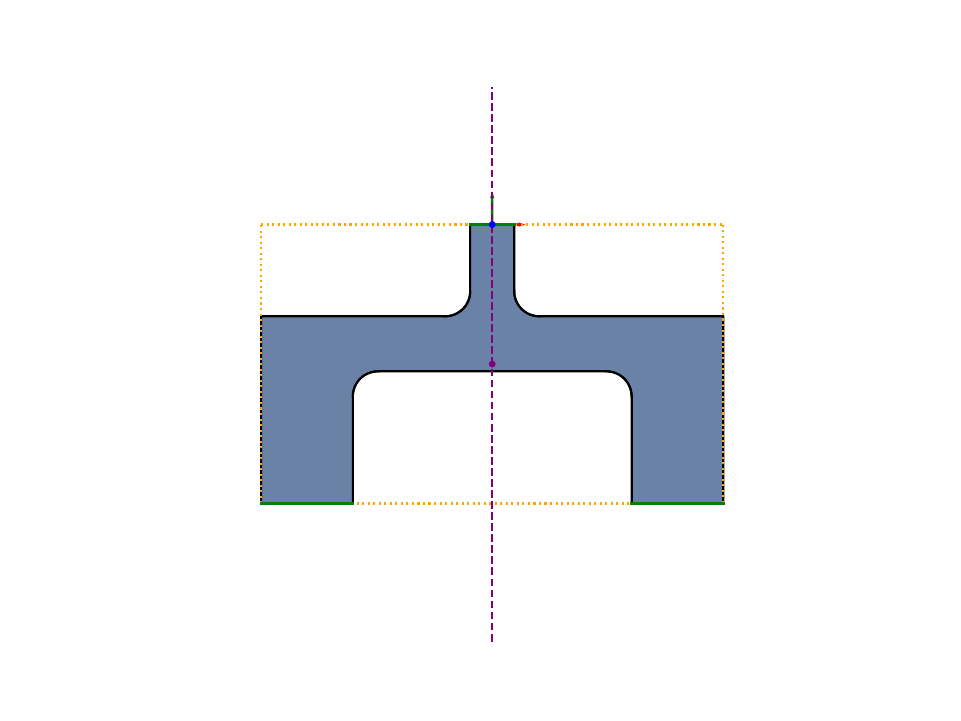} &    
  \includegraphics[width=0.32\textwidth,clip,trim={0.2\wd0 0.2\wd0 0.2\wd0 0.2\wd0}]{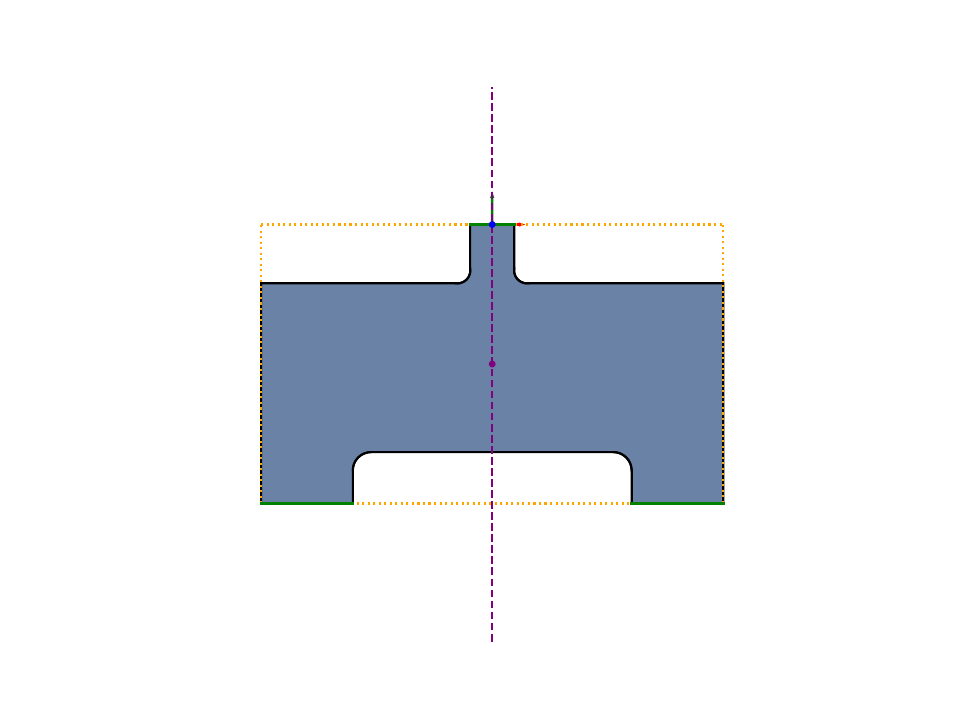} \\
  Area 0.2 & Area 0.4 & Area 0.8\\
  \end{tabular}
  \caption{Profiles generated when three different profile areas requested.  When an area of 0.2 model units is requested, the generated profile contains thin regions.  Increasing the requested area allows shapes without this thin region.}
  \label{fig:geom_prompt_area}
\end{figure}

\begin{figure}[ht]
  \centering
  \sbox0{\includegraphics{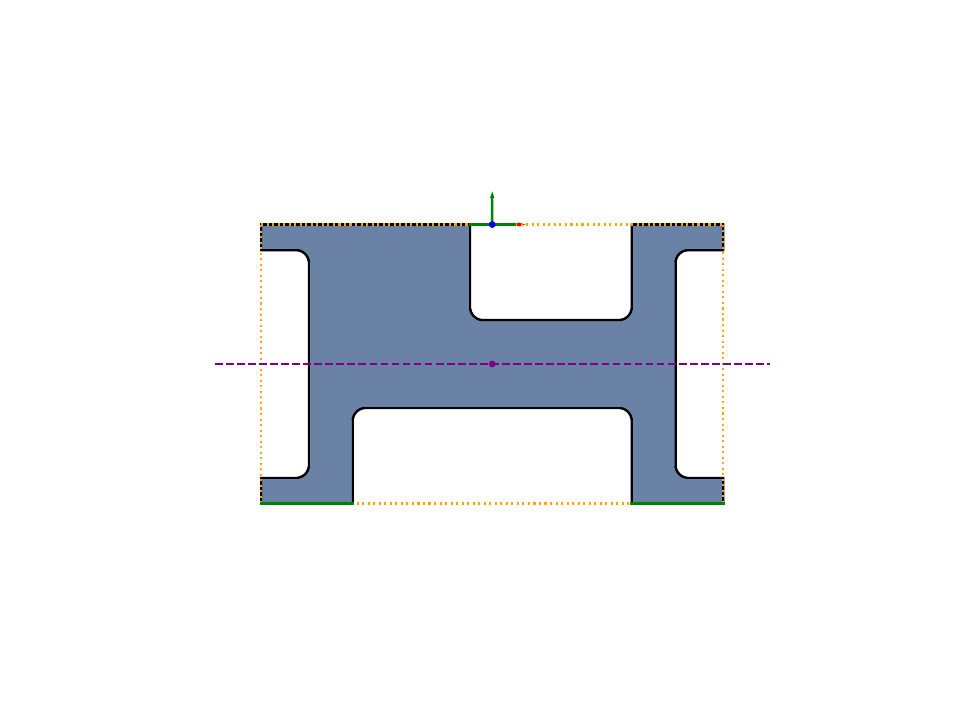}}
  \begin{tabular}{ccc}
  \includegraphics[width=0.32\textwidth,clip,trim={0.2\wd0 0.2\wd0 0.2\wd0 0.2\wd0}]{figures/geometric_prompt/Symmetry/787_first_generation_00.pdf} &
  \includegraphics[width=0.32\textwidth,clip,trim={0.2\wd0 0.2\wd0 0.2\wd0 0.2\wd0}]{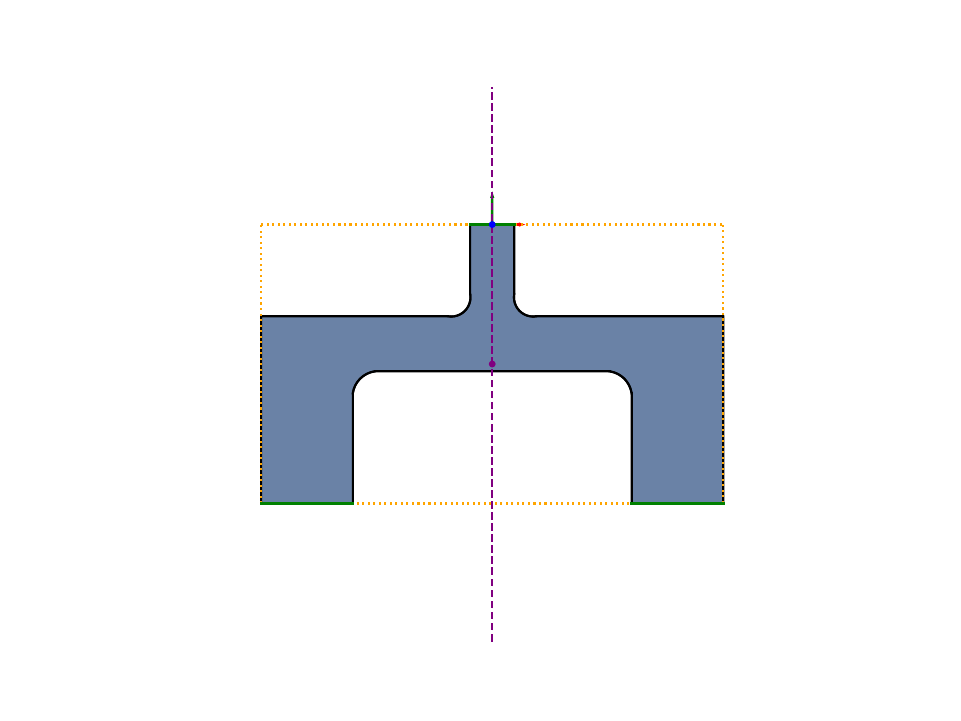} &    
  \includegraphics[width=0.32\textwidth,clip,trim={0.2\wd0 0.2\wd0 0.2\wd0 0.2\wd0}]{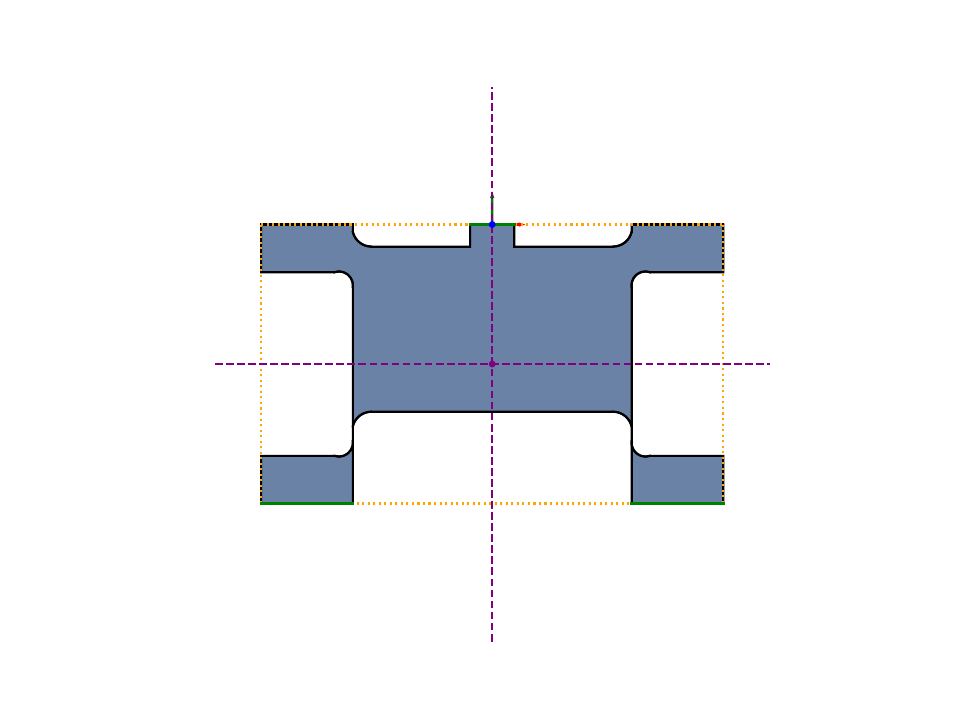} \\
  Horizontal symmetry line & Vertical symmetry line & Horizontal and vertical symmetry lines\\
  \end{tabular}
  \caption{Requesting different symmetry lines in the geometric prompt.}
  \label{fig:geom_prompt_symmetry}
\end{figure}

\begin{figure}[ht]
  \centering
  \sbox0{\includegraphics{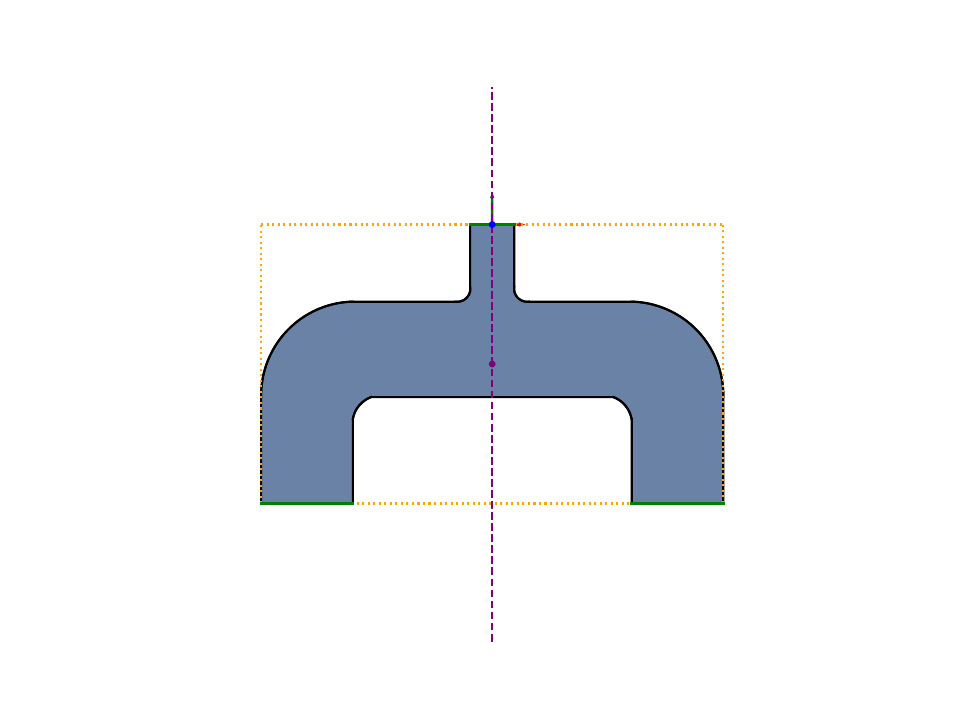}}
  \begin{tabular}{ccc}
  \includegraphics[width=0.32\textwidth,clip,trim={0.2\wd0 0.2\wd0 0.2\wd0 0.2\wd0}]{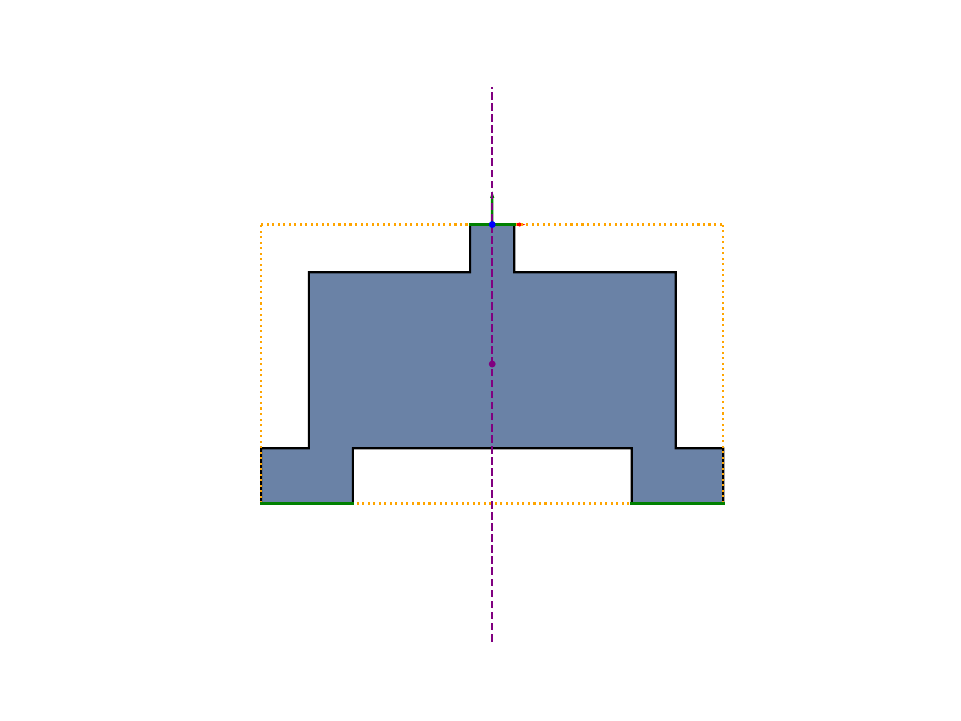} &
  \includegraphics[width=0.32\textwidth,clip,trim={0.2\wd0 0.2\wd0 0.2\wd0 0.2\wd0}]{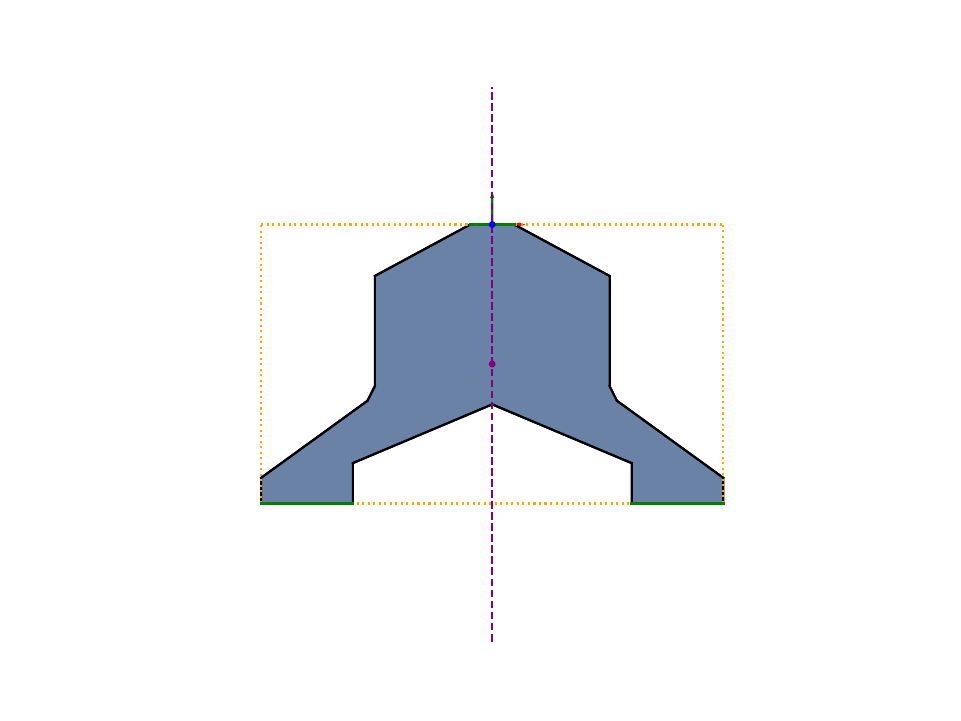} &    
  \includegraphics[width=0.32\textwidth,clip,trim={0.2\wd0 0.2\wd0 0.2\wd0 0.2\wd0}]{figures/geometric_prompt/SmoothFraction/797_smooth_fraction_0.8_first_generation_07.pdf} \\
 Tangent continuous vertices 0.0 & Tangent continuous vertices 0.3 & Tangent continuous vertices 0.8\\
  \end{tabular}
  \caption{Profiles obtained by varying the requested fraction of tangent continuous vertices.}
  \label{fig:geom_prompt_smooth_fraction}
\end{figure}

\begin{figure}[ht]
  \centering
  \sbox0{\includegraphics{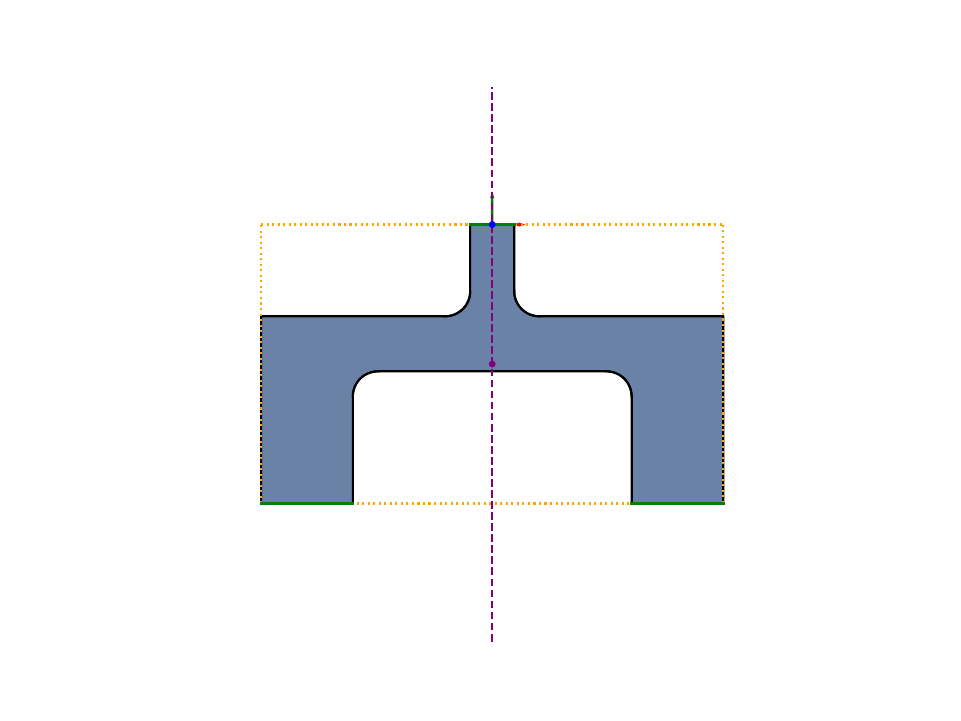}}
  \begin{tabular}{ccc}
  \includegraphics[width=0.32\textwidth,clip,trim={0.2\wd0 0.2\wd0 0.2\wd0 0.2\wd0}]{figures/geometric_prompt/NumEdges/780-first_generation_01.pdf} &
  \includegraphics[width=0.32\textwidth,clip,trim={0.2\wd0 0.2\wd0 0.2\wd0 0.2\wd0}]{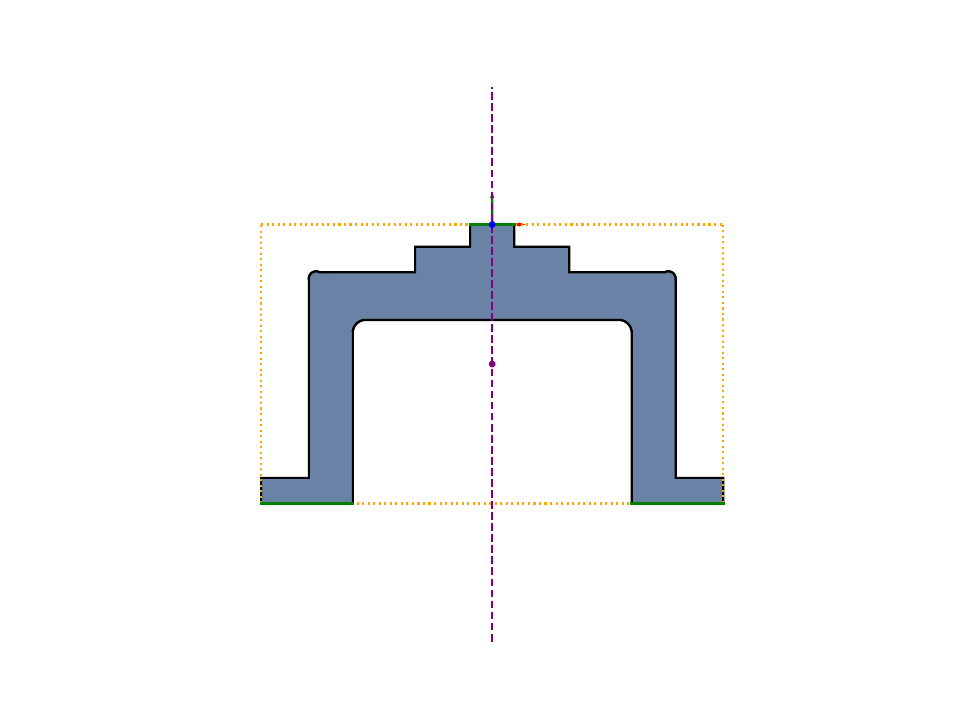} &    
  \includegraphics[width=0.32\textwidth,clip,trim={0.2\wd0 0.2\wd0 0.2\wd0 0.2\wd0}]{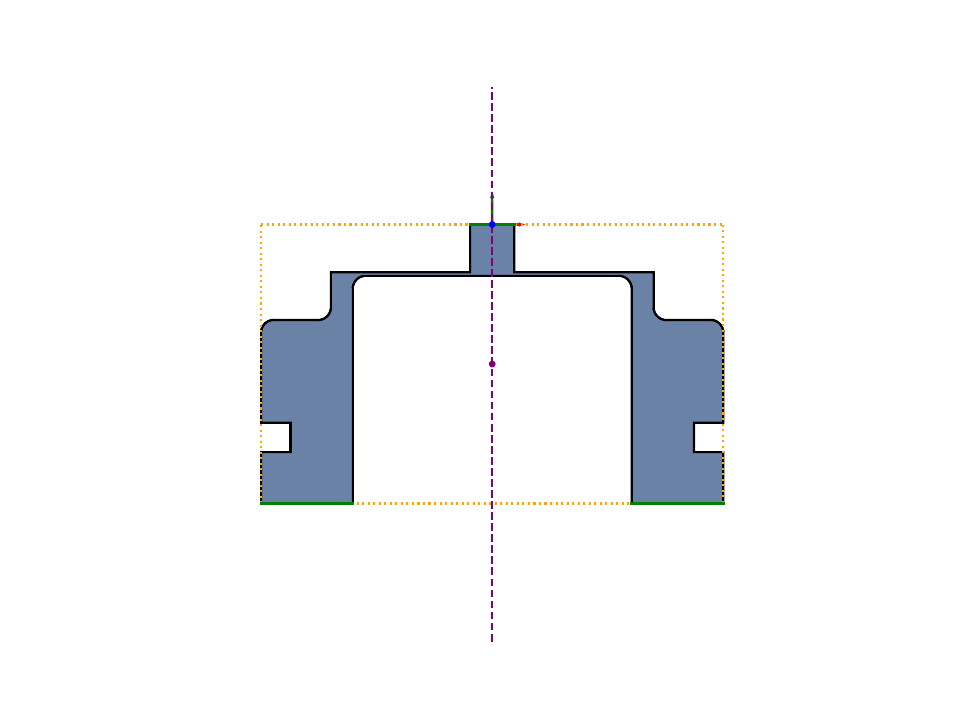} \\
Number of edges 16 & Number of edges 26 & Number of edges 32\\
  \end{tabular}
  \caption{Controlling shape complexity by requesting different numbers of edges in the geometric prompt.}
  \label{fig:geom_prompt_num_edges}
\end{figure}

\begin{figure}[ht]
  \centering
  \sbox0{\includegraphics{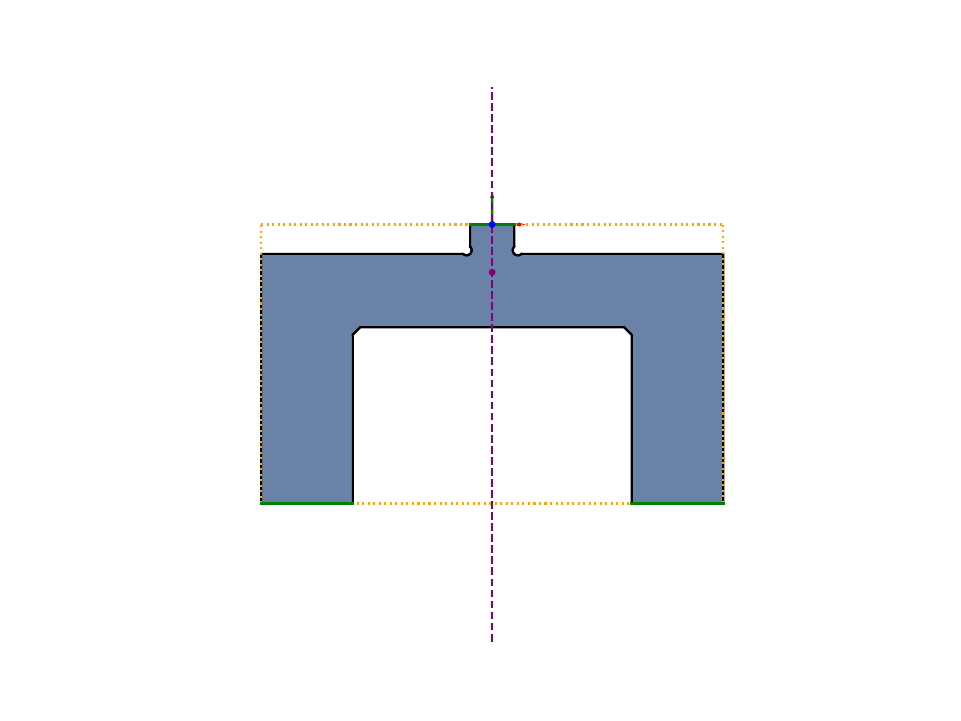}}
  \begin{tabular}{ccc}
  \includegraphics[width=0.32\textwidth,clip,trim={0.2\wd0 0.2\wd0 0.2\wd0 0.2\wd0}]{figures/geometric_prompt/CoG/800_cog_0.4_first_generation_00.pdf} &
  \includegraphics[width=0.32\textwidth,clip,trim={0.2\wd0 0.2\wd0 0.2\wd0 0.2\wd0}]{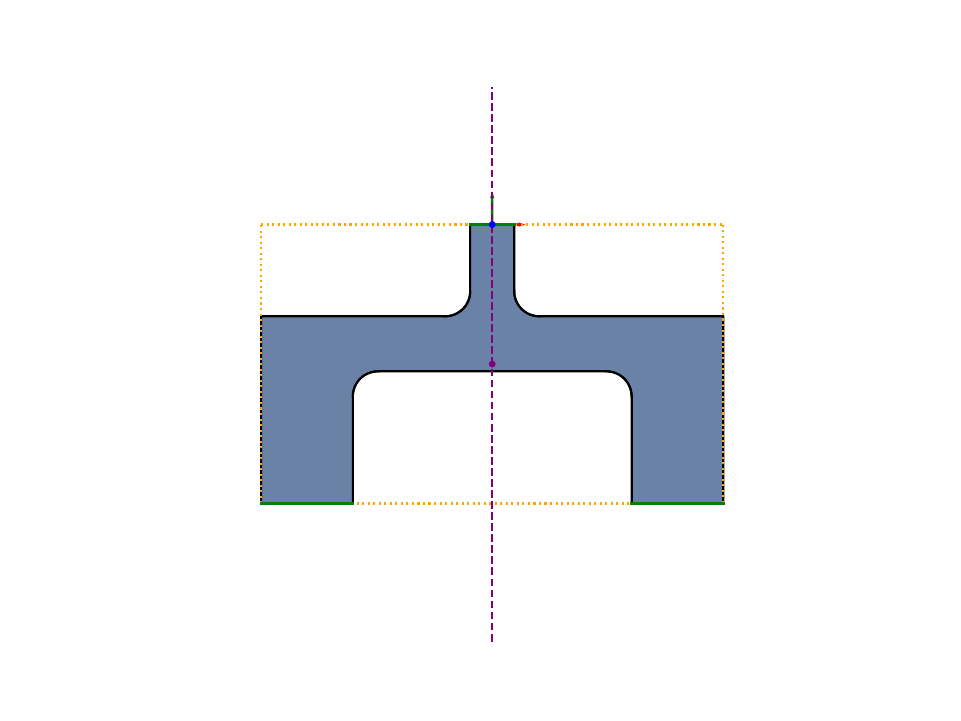} &    
  \includegraphics[width=0.32\textwidth,clip,trim={0.2\wd0 0.2\wd0 0.2\wd0 0.2\wd0}]{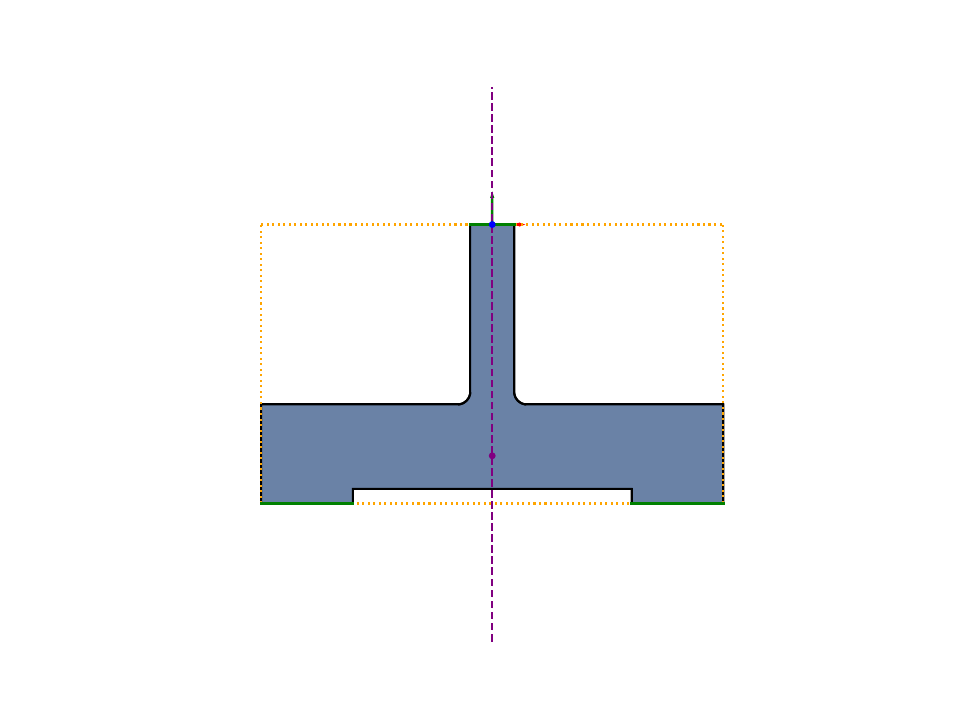} \\
Center of gravity high & Center of gravity middle  & Center of gravity low\\
  \end{tabular}
  \caption{The effect of raising and lowering the requested center of gravity.  The center of gravity is marked as a purple point.}
  \label{fig:geom_prompt_cog}
\end{figure}

\FloatBarrier

\begin{figure}[ht]
  \centering
  \sbox0{\includegraphics{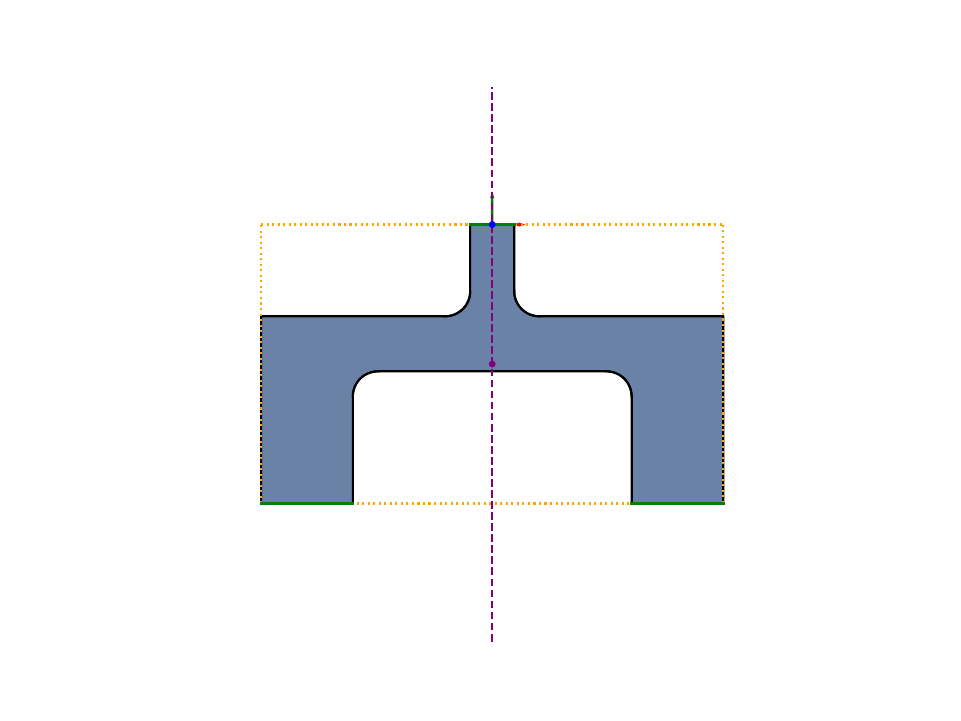}}
  \begin{tabular}{ccc}
  \includegraphics[width=0.32\textwidth,clip,trim={0.2\wd0 0.2\wd0 0.2\wd0 0.2\wd0}]{figures/geometric_prompt/TopWidth/780_-_standard_first_generation_00.pdf} &
  \includegraphics[width=0.32\textwidth,clip,trim={0.2\wd0 0.2\wd0 0.2\wd0 0.2\wd0}]{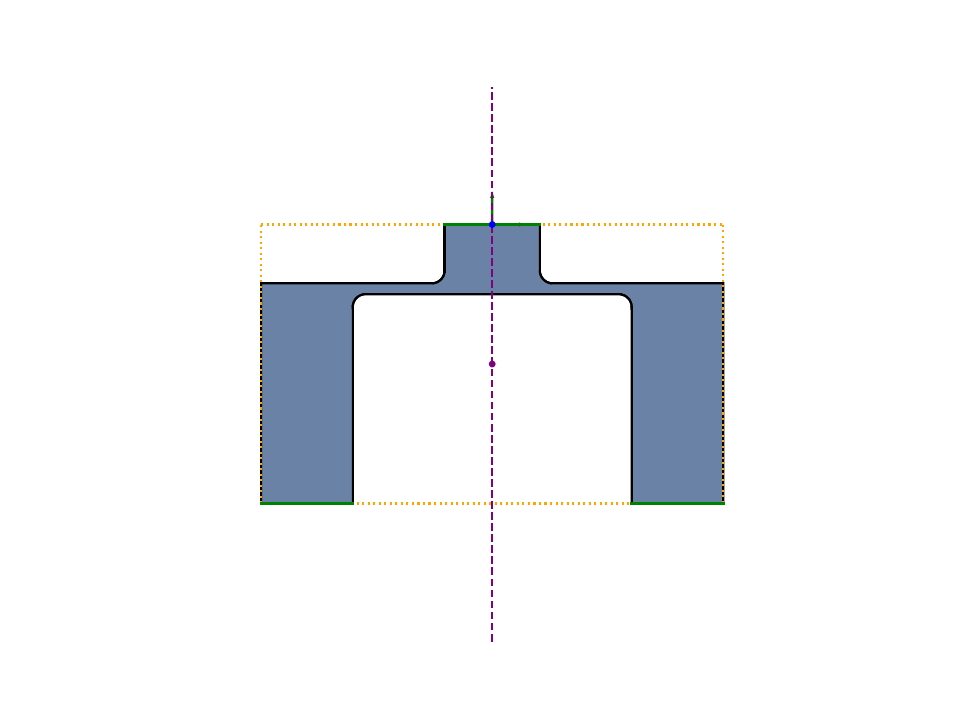} &    
  \includegraphics[width=0.32\textwidth,clip,trim={0.2\wd0 0.2\wd0 0.2\wd0 0.2\wd0}]{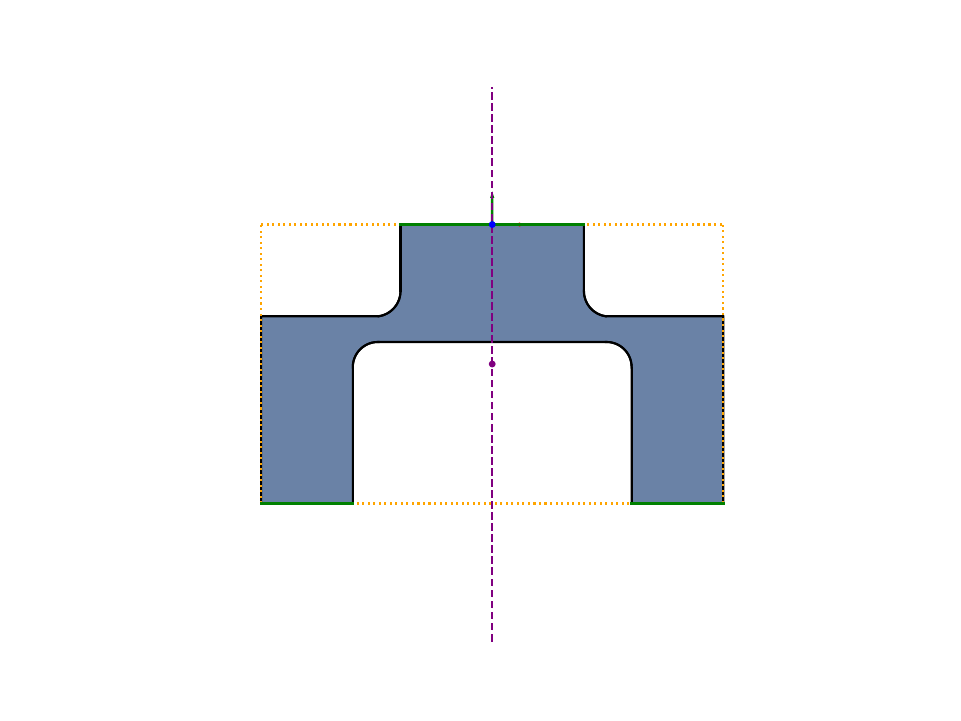} \\
Top length 0.1 & Top length 0.2 & Top length 0.4\\
  \end{tabular}
  \caption{The effect of varying the length of the bounded line segment (green) at the top of the figure.  As this length is changed, the top part of the profile widens in a natural way.}
  \label{fig:geom_prompt_top_width}
\end{figure}

Designers often want control over the mechanical strength and weight of the parts being modeled.  We observe that changing the requested area of the generated profile provides some control over thickness and consequently allows mechanical strength to be increased at the expense of increased mass.  The generated shape with three different requested areas is shown in Figure \ref{fig:geom_prompt_area}.

The ability to make designs symmetrical is also important for designers.  In many cases, designers start by drawing symmetry lines  and then use constraints to keep various profile curves symmetric around these lines.   In Figure \ref{fig:geom_prompt_symmetry} we show how including horizontal and vertical symmetry lines in the geometric prompt effects the resulting generation.  Notice that the bounded line segments  which the profile is requested to match (green lines) exhibit symmetry consistent with a vertical symmetry line only.  When a horizontal symmetry line is requested, the network is unable to match both the bounded lines and the requested symmetry.  Figure \ref{fig:geom_prompt_symmetry_construction_sequence} of Appendix \ref{appendix:generated_constrction_sequence_examples} shows the detailed construction sequence when both horizontal and vertical symmetries are included.  From this sequence we can see how symmetries in the final shape are derived by mirroring points over the symmetry lines provided.

A useful way to provide some high level control over the style of the shape is by requesting sharp corners or tangent continuous vertices. In Figure \ref{fig:geom_prompt_smooth_fraction} we see how increasing the requested fraction of tangent continuous vertices allows the shape to be varied from a rectilinear form (left) to the smoothly curved shape shown on the right.  

The complexity of the generated profile can be adjusted by specifying the number of edges in the geometric prompt. Figure \ref{fig:geom_prompt_num_edges} illustrates how increasing the number of edges produces more intricate shapes. However, we find that when higher complexity is requested, the additional details are often poorly defined. In practice, reducing the requested complexity can help eliminate these unnecessary or ambiguous features.  

In cases where the desired shape is ambiguous, additional control can be exerted by specifying the desired center of gravity for the profile.  In Figure  \ref{fig:geom_prompt_cog} we show how the raising and lowering the center of gravity along the symmetry line changes the final shape.

Finally, in Figure \ref{fig:geom_prompt_top_width} we demonstrate that changing the bounded line segments provided in the prompt, while keeping all other properties at the values defined in Table \ref{tab:geom_prompt_standard}, results in sensible shape changes. 

\subsection{Evaluation metrics}
Evaluation metrics for 2d parametric profile generation can be broadly divided into two categories: \textit{validity metrics}, which assess whether a generated profile sequence is syntactically correct and geometrically sound, and \textit{prompt satisfaction metrics}, which evaluate how well the generated profile shape adheres to the constraints and properties specified in the input geometric prompt.

\subsubsection{Validity metrics}
Validity metrics measure whether the generated profiles conform to both the syntactic requirements of the DSL and the implicit geometric expectations of a well-formed profile shape. These metrics include:

\noindent\textbf{Syntactic validity}: Whether a generated sequence can be successfully detokenized under the strict syntactic rules of the DSL.

\noindent\textbf{No self-intersection}: Whether the resulting profile is free of self-intersections.

\noindent\textbf{No short edges}: Whether all edges exceed the minimum length defined by the quantization bin size.

These are typically reported as boolean indicators, aggregated as the overall fraction of valid profile generations.

\subsubsection{Prompt satisfaction metrics}
The degree to which a generated profile adheres to the geometric prompt can then be quantitatively evaluated, enabling direct and objective assessment of prompt satisfaction. Some of the prompt satisfaction metrics include:

\noindent\textbf{Area}: measured by the difference between the area prescribed in the geometric prompt and that of the generated profile.

\noindent\textbf{Line segments}: for each line segment specified in the prompt, the metric is computed based on the presence of profile line segments that are collinear, overlapping, and equal in length. The distance between the end points of the requested and generated line segments is also recorded.

\noindent\textbf{Center-of-gravity}: measured by the distance between the center of gravity defined in the geometric prompt and that of the generated profile.

\noindent\textbf{Holes}: measured by the distances between the hole centers defined in the geometric prompt and that of the generated profile.

\noindent\textbf{Symmetry lines}: measured by the intersection over union  (IoU) of the profile and its reflection across the symmetry line, averaged over all symmetry lines in the prompts.

\noindent\textbf{Outer bounding box}: measured by the intersection over union between the outer bounding box defined in the geometric prompt and that of the generated profile.

\noindent\textbf{Fraction of tangent continuous vertices}: measured by the difference between the fraction of tangent continuous vertices prescribed in the geometric prompt and that of the generated profile.

\subsection{Quantitative results}
\label{results:quantititive}
Table~\ref{tab:metrics_comparison} presents a comparative analysis of evaluation metrics across five models: a baseline model trained without construction sequences; a construction steps model trained with construction sequences with the same hyperparameters as the baseline model; and three RL finetuned variants of the construction steps model. These aligned variants, ReMax, GRPO, and RLOO, are optimized using reward functions defined in Section~\ref{subsection:reward_design}. All models are evaluated with greedy sampling of top k equals to one. 

The introduction of construction sequences leads to substantial improvements over the baseline across all validity metrics and most of the prompt satisfaction metrics. Syntactic validity increases from 88.1\% to 94.0\%, while the proportion of non-self-intersecting profiles increases from 81.9\% to 84\% and compliance with minimum edge length increased from 88.2\% to 94.3\%. Among the prompt satisfaction metrics, the most notable gains are observed in line segment adherence and mirror symmetry. These results confirm that integrating construction sequences alone significantly enhances both structural validity and alignment with geometric constraints.

Further gains are realized through RL-based alignment. The aligned variants consistently outperform the unaligned construction steps model across syntactic and geometric validity metrics, including self-intersection avoidance and minimum edge length compliance. For instance, both RLOO and GRPO achieve more than 6\% reduction in generating self-intersecting geometries. Notably, although the reward functions are explicitly designed to optimize geometric validity, we observe consistent and often substantial gains across a broad set of geometric prompting metrics, including area accuracy, bounding box alignment, symmetry, and hole placement. This suggests that structural improvements induced by alignment not only satisfy low-level constraints but also enhance higher-level geometric properties, even when these are not directly incentivized during optimization.

\begin{figure}[!htb]
    \centering
    \includegraphics[width=0.8\textwidth]{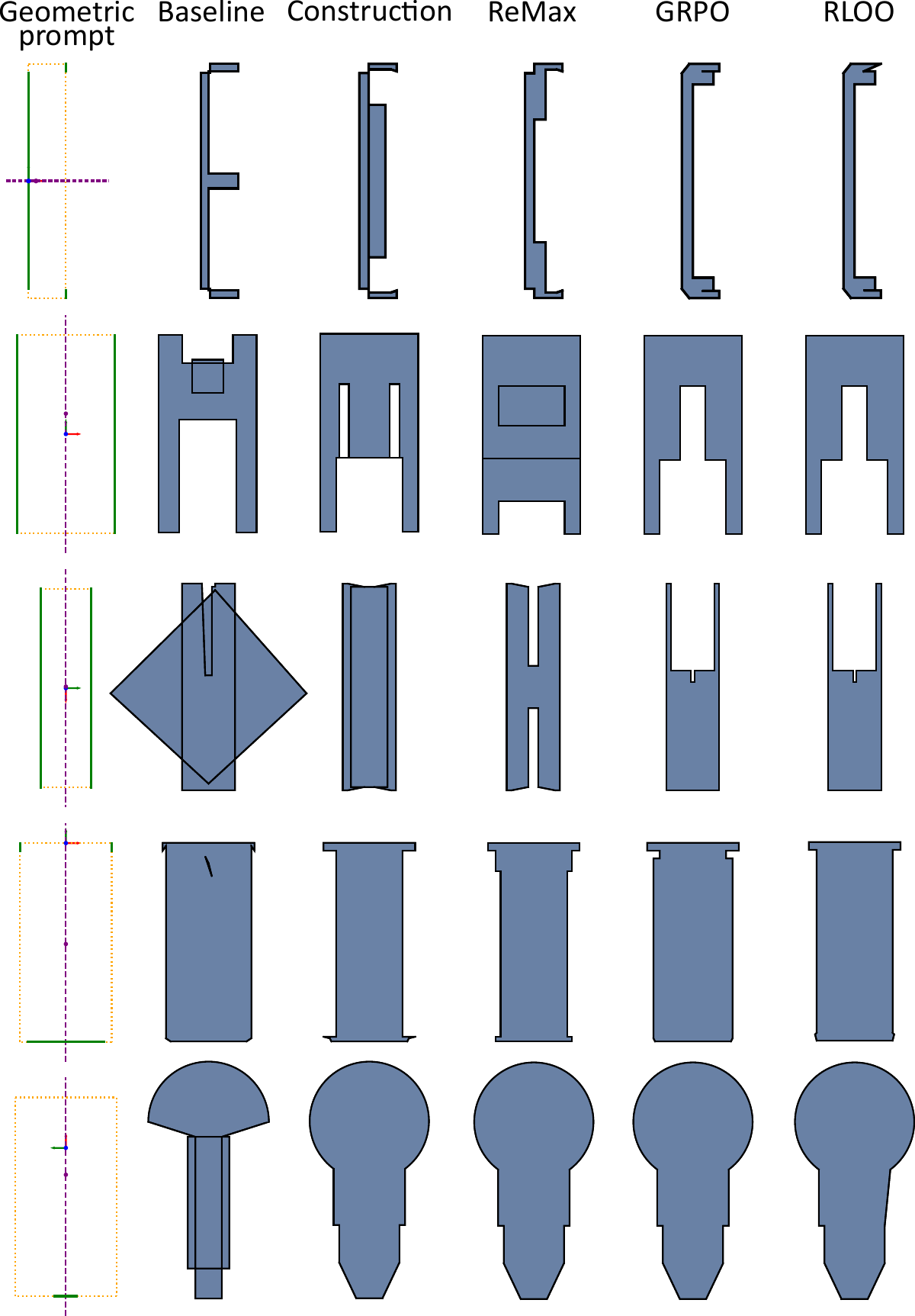}
    \caption{Visual comparison of profiles generated with the base model without construction sequences, the construction sequences model and the aligned models.}
    \label{fig:qualitative_results}
\end{figure}


\begin{table}[!htb]
    \centering
    \caption{Comparison of key metrics among different models}
    \label{tab:metrics_comparison}
    \begin{tabular}{@{} l c c c c c @{}}
        \toprule
        \textbf{Metrics} &
        \makecell{\textbf{Baseline} \\ \textbf{model}} &
        \makecell{\textbf{Construction} \\ \textbf{steps model}} &
        \makecell{\textbf{Construction} \\ \textbf{steps model}\\ \textbf{(ReMax)}} &
        \makecell{\textbf{Construction} \\ \textbf{steps model}\\ \textbf{(GRPO)}} &
        \makecell{\textbf{Construction} \\ \textbf{steps model}\\ \textbf{(RLOO)}} \\
        \midrule
        Syntactic validity ($\uparrow$) & 0.881 & 0.940 & 0.945 & 0.975 & \textbf{0.976}\\ 
        \midrule
        No self-intersection ($\uparrow$) & 0.819 & 0.840 & 0.853 & 0.903 & \textbf{0.905}\\
        \midrule
        No short edges ($\uparrow$) & 0.882 & 0.943 & 0.948 & 0.976 & \textbf{0.978} \\
        \midrule
        Difference in area ($\downarrow$) & 0.253 & 0.238 & 0.210 & 0.170 & \textbf{0.162}\\
        \midrule
        Line segment dist ($\downarrow$) & 0.00313 & 0.00152 & \textbf{0.00043} & 0.00082 & 0.00111\\
        \midrule
        Line segment ratio ($\uparrow$) & 0.963 & \textbf{0.983} & 0.981 & 0.978 & 0.976\\
        \midrule
        \makecell[l]{Center-of-gravity \\distance ($\downarrow$)} & \textbf{0.0247} & 0.0267 & 0.0285 & 0.0254 & 0.0265\\
        \midrule
        Hole center dist ($\downarrow$) & 0.0255 & 0.0318 & \textbf{0.0153} & 0.0232 & 0.0402 \\
        \midrule
        Mirror IoU ($\uparrow$) & 0.818 & 0.859 & 0.865 & \textbf{0.886} & \textbf{0.886}\\
        \midrule
        \makecell[l]{Outer bounding\\box IoU ($\uparrow$)} & \textbf{0.990} & 0.984 & 0.981 & 0.985 & 0.983 \\
        \midrule
        \makecell[l]{Tangent continuous\\ vertices  
        difference ($\downarrow$)} & 0.0907 & \textbf{0.0786} & 0.0799 & 0.0814 & 0.0816\\
        \bottomrule
    \end{tabular}
\end{table}

\section{Qualitative results}
\label{appendix:qualitative_results}
Figure~\ref{fig:qualitative_results} presents qualitative comparisons of profiles generated by different models. The first column illustrates the geometric prompt used to drive the generation process, while the other columns show the generated profile geometry.  These results demonstrate that in cases where the baseline model fails to adhere to the complex structural constraints encoded in the prompts, the construction steps model and its aligned variants can successfully produce geometrically valid  generations faithful to the design specification. 
Further qualitative results are shown in Appendix \ref{appendix:shape_control_from_geometric_prompt} and \ref{appendix:generated_constrction_sequence_examples}.  These demonstrate how the geometric prompt can be used to control the shape of the generated profile and include examples of the full construction sequences and the family of shapes which can be obtained from them by varying the driving parameters.
\section{Conclusion}
In this work, we introduced a new sequence representation for CAD generation, which constructs profiles using a sequence of simple geometric construction steps. Adding these construction steps between the designers input and the final shape improves the generation quality, and promotes adherence to design requirements. Furthermore, we showed that reinforcement learning, guided by reward functions that penalize self-intersections, achieves consistent improvements across a range of metrics, including those not explicitly targeted. As the generated sequences can be replayed with floating point precision, they overcome the accuracy limitations of previous methods.  Additionally, by reducing the degrees of freedom into a small set of parameters, the resulting shapes can be manipulated parametrically, much like edits in a parametric CAD system.


\bibliography{refs.bib}
\bibliographystyle{unsrt}

\appendix

\section{Domain specific language and tokenization for construction sequences}
\label{appendix:tokenization}

\subsection{Geometry}
\label{appendix:construction_step_geometry}
The domain specific language used to represent geometric prompts, construction sequences and the final profiles, share a common vocabulary for geometric entities.  Scalar values, points and infinite lines are represented by single tokens as shown in Table \ref{tab:single_token_entities}.   More complex entities are then constructed using a combination of special tokens denoting the entity type, followed by primitive tokens representing the geometry as shown in Table \ref{tab:multi_token_entities}.

\begin{table}[!ht]
    \centering
    \caption{Geometry and scalar values represented by single tokens}
    \label{tab:single_token_entities}
    \renewcommand{\arraystretch}{0.5}
    \begin{tabular}{@{}p{0.2\textwidth} p{0.75\textwidth}}
        \toprule
        \textbf{Token} & \textbf{Explanation}  \\
        \midrule

        \texttt{Length} 
        &
        Signed lengths in the range $[-1, 1]$ are quantized into 127 bins.  An odd number of bins is deliberate chosen so that the values of $-1.0$, $0.0$ and $1.0$ can be recovered exactly when the quantized values are converted back to floating point numbers.
        \\

        \midrule
        \texttt{Angle} 
        &
        Angles in the range $[0, 2\pi)$ are quantized into 121 bins. Care is taken to ensure the that the angles  $0.0$ and $2\pi$ map to the center of the first bin.  As 120 has prime factors $2$, $3$ and $5$, this quantization strategy allows common angles like $0.0$, $\pi$, $\pi/2$ and $\pi/3$ to be exactly reconstructed.
        \\

        \midrule
        \texttt{Point} 
        &
        Points are quantized onto a $127\times127$ grid.  An odd number of grid cells in each direction is chosen to ensure that the center point in the domain can be exactly reconstructed when the quantized representation is de-quantized.
        \\

        \midrule
        \texttt{Infline} 
        &
        Directed infinite lines are represented in hessian normal form.  The angle between the line and the $x$-axis is encoded into $121$ bins and the signed distance to the origin is encoded into $127$ bins.  Infinite lines are then represented as single tokens with $121\times127$ possible values.
        \\

        \midrule
        \texttt{Area} 
        &
        Area values are used in the geometric prompt to control profile thickness.  After normalization, the profiles' areas  will be in the range $[0, 1]$ and can be  quantized into 127 bins. 
        \\

        \midrule
        \texttt{Complexity} 
        &
        The number of curves in the profile is used as a measure of complexity.   Complexity tokens have 127 possible values with the last token value used for profiles containing more than 127 curves. 
        \\

        \midrule
        \texttt{NumLoops} 
        &
        The number of loops in the profile is represented by a single token with 127 possible values.
        \\

        \midrule
        \texttt{SmoothVertices} 
        &
        The fraction of vertices which are tangent continuous is quantized into 127 bins.
        \\

        \bottomrule
    \end{tabular}
\end{table}


\newpage

\begin{table}[!ht]
    \centering
    \caption{Geometric entities defined by a combination of multiple tokens.  Each multi-token entity starts with a special token denoting the entity type followed by a list of tokens to define the required geometry.}
    \label{tab:multi_token_entities}
    \renewcommand{\arraystretch}{0.5}
    \begin{tabular}{@{}p{0.35\textwidth} p{0.65\textwidth}}
        \toprule
        \textbf{Tokens} & \textbf{Explanation}  \\

        \midrule
        \begin{tabular}[t]{@{\hspace{0em}}p{0.3\linewidth} p{0.6\linewidth}}
            \texttt{BoundLine} \\
            \hspace*{1em}\texttt{Point} &// \emph{Start point} \\
            \hspace*{1em}\texttt{Point} &//  \emph{End point}
        \end{tabular}

        &
        Bounded line segments start with a \texttt{BoundedLine}  token followed by two point tokens representing the start and the end point of the line segment.
        \\

        \midrule
        \begin{tabular}[t]{@{\hspace{0em}}p{0.3\linewidth} p{0.6\linewidth}}
          \texttt{Arc} & \\
          \hspace*{1em}\texttt{Point} & // \emph{Start point} \\
          \hspace*{1em}\texttt{Point} & // \emph{Mid point} \\
          \hspace*{1em}\texttt{Point} & // \emph{End point}
        \end{tabular}

        &
        Arcs are represented with an \texttt{Arc} type token following three point tokens representing the start, mid and end points of the arc segment.
        \\

        \midrule
        \begin{tabular}[t]{@{\hspace{0em}}p{0.3\linewidth} p{0.6\linewidth}}
          \texttt{Circle} & \\
          \hspace*{1em}\texttt{Point}  & // \emph{Center point} \\
          \hspace*{1em}\texttt{Length} & // \emph{Radius} \\
          \hspace*{1em}\texttt{IsCCW}  & // \emph{Counter-clockwise}
        \end{tabular}

        &
        Oriented circles are represented with a point token for the center point and a length token for the circle's radius.  An additional flag indicates whether the circle is oriented clockwise or counter-clockwise.
        \\

        \midrule
        \begin{tabular}[t]{@{\hspace{0em}}p{0.3\linewidth} p{0.6\linewidth}}
          \texttt{BoundingBox} \\
            \hspace*{1em}\texttt{Point} &// \emph{Min point} \\
            \hspace*{1em}\texttt{Point} &// \emph{Max point}
        \end{tabular}

        &
        A bounding box is encoded by two points at the min and max corners.
        \\

        \midrule
        \begin{tabular}[t]{@{\hspace{0em}}p{0.3\linewidth} p{0.6\linewidth}}
            \texttt{DimensionDatum} \\
            \hspace*{1em}\texttt{Point} &// \emph{Datum point} \\
            \hspace*{1em}\texttt{Infline} &//  \emph{x-axis} \\
            \hspace*{1em}\texttt{Infline} &//  \emph{y-axis}
        \end{tabular}

        &
        The dimension datum defines a fixed reference point from which other geometry can be constructed. It consists of a datum point with two infinite lines passing through it, aligned with the x and y coordinate system axes.
        \\

        \midrule
        \begin{tabular}[t]{@{\hspace{0em}}p{0.3\linewidth} p{0.6\linewidth}}
          \texttt{BoltHole} & \\
          \hspace*{1em}\texttt{Point}  & // \emph{Center point} \\
          \hspace*{1em}\texttt{Length} & // \emph{Radius} \\
          \hspace*{1em}\texttt{Length} & // \emph{Clearance disk}
        \end{tabular}

        &
        Bolt holes are common in mechanical parts.  They are used in the geometric prompt to define both the hole's position and the required clearance around it.   The tokenization contains the center point, a radius representing the size of the hole's aperture and a clearance disk representing material which should be kept in place surrounding the hole to prevent tear-out of the bolt.
        \\
        
        \bottomrule
    \end{tabular}
\end{table}


\subsection{Geometric prompt}
\begin{figure}[H]
    \centering
    \includegraphics[width=\textwidth]{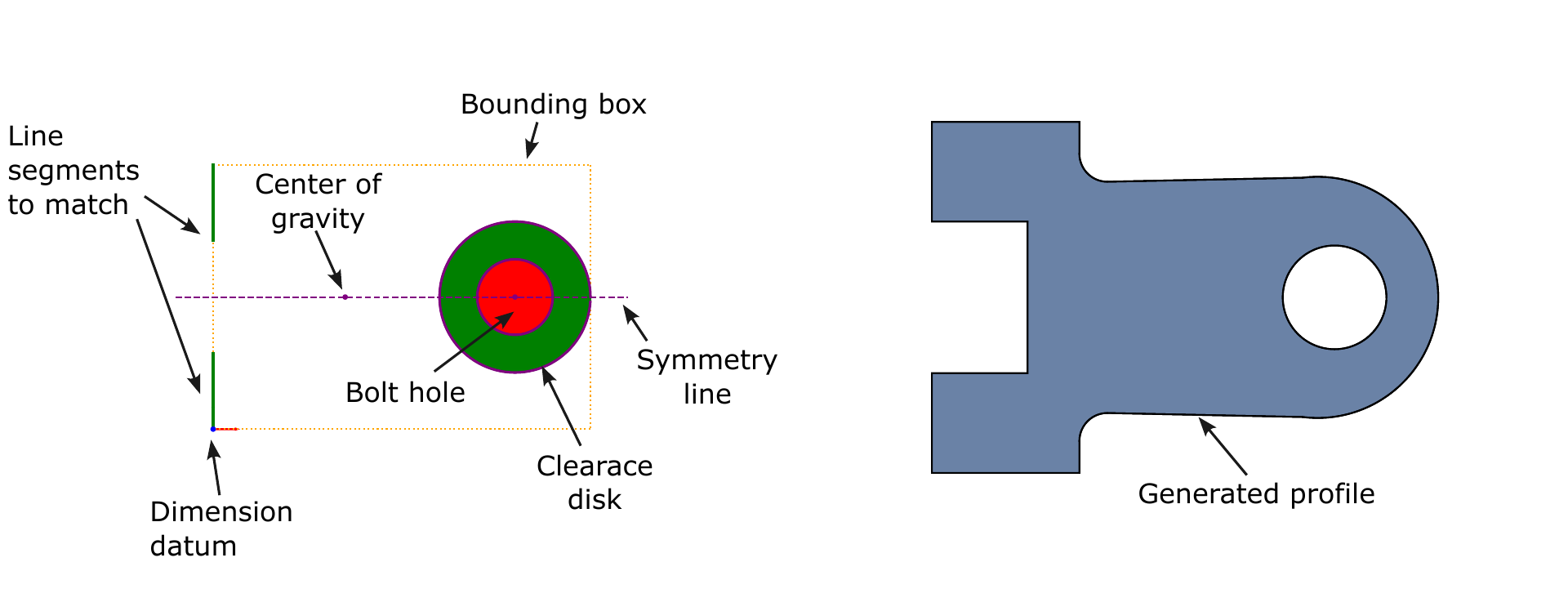}
    \caption{The geometric elements used in a geometric prompt and the resulting generated profile.}
    \label{fig:prompt_and_profile}
\end{figure}

The generation is guided by geometric prompts, which specify the geometry the generated profile should match, along with other important properties such as profile area and center of gravity. Appendix \ref{appendix:appendix_geometric_prompts} describes the rationale for each component in the geometric prompts in more detail and Appendix \ref{appendix:shape_control_from_geometric_prompt} shows some examples of how the prompt can be used to control the shape.  The tokenization of the prompts is shown in Table \ref{tab:geometric_prompt} which includes lower level entities from Tables \ref{tab:single_token_entities} and \ref{tab:multi_token_entities}.  An illustration of a geometric prompt and the generated profile is show in Figure\ref{fig:prompt_and_profile}.

\newlength{\colOneWidth}
\newlength{\colTwoWidth}
\setlength{\colOneWidth}{0.2\linewidth}
\setlength{\colTwoWidth}{0.2\linewidth}

\begin{table}[!ht]
    \centering
    \caption{Geometric prompt entities}
    \label{tab:geometric_prompt}
    \renewcommand{\arraystretch}{0.5}
    \begin{tabular}{@{}p{0.5\textwidth} p{0.5\textwidth}}
        \toprule
        \textbf{Tokens} & \textbf{Explanation}  \\
        \midrule


        \begin{tabular}[t]{@{\hspace{0em}}p{\colOneWidth} p{\colTwoWidth}}
            \texttt{StartOfPrompt} 
        \end{tabular}

        &
        A special token denoting the start of the geometric prompt.
        \\
        \\


        \hspace*{1em}\begin{tabular}[t]{@{\hspace{0em}}p{\colOneWidth} p{\colTwoWidth}}
            \texttt{DimensionDatum} 
        \end{tabular}

        &
       The dimension datum.  The construction steps will build up the profile geometry relative to this fixed datum point.
        \\
        \\


        \hspace*{1em}\begin{tabular}[t]{@{\hspace{0em}}p{\colOneWidth} p{\colTwoWidth}}
            \texttt{BoundingBox} 
        \end{tabular}

        &
        The desired bounding box for the profile being generated.
        \\
        \\
        

        \hspace*{1em}\begin{tabular}[t]{@{\hspace{0em}}p{\colOneWidth} p{\colTwoWidth}}
            \texttt{Area} \\
        \end{tabular}

        &
        The desired area of the profile being generated.
        \\


        \hspace*{1em}\begin{tabular}[t]{@{\hspace{0em}}p{\colOneWidth} p{\colTwoWidth}}
            \texttt{Complexity} 
        \end{tabular}

        &
        The approximate number of curves we would like the generated profile to contain. 
        \\
        \\
        

        \hspace*{1em}\begin{tabular}[t]{@{\hspace{0em}}p{\colOneWidth} p{\colTwoWidth}}
            \texttt{NumLoops} 
        \end{tabular}

        &
        The desired number of loops for the generated profile.
        \\
        \\


        \hspace*{1em}\begin{tabular}[t]{@{\hspace{0em}}p{\colOneWidth} p{\colTwoWidth}}
            \texttt{SmoothVertices} 
        \end{tabular}

        &
        The desired fraction of vertices which we would like to be tangent continuous.
        \\
        \\


        \hspace*{1em}\begin{tabular}[t]{@{\hspace{0em}}p{\colOneWidth} p{\colTwoWidth}}
            \texttt{CenterOfGravity} \\
            \hspace*{1em}\texttt{Point}  
        \end{tabular}

        &
        The desired center of gravity of the profile.
        \\
        \\
        

        \hspace*{1em}\begin{tabular}[t]{@{\hspace{0em}}p{\colOneWidth} p{\colTwoWidth}}
            \texttt{SymmetryLines} \\
            \hspace*{1em}\texttt{ListOf(Infline)}  
        \end{tabular}

        &
        Lines defining the mirror symmetries we would like the generated profile to have.
        \\
        \\


        \hspace*{1em}\begin{tabular}[t]{@{\hspace{0em}}p{\colOneWidth} p{\colTwoWidth}}
            \texttt{BoundLines} \\
            \hspace*{1em}\texttt{ListOf(BoundLine)}  
        \end{tabular}

        &
         A list of line segments which the the generated profile should include.
        \\
        \\





        \hspace*{1em}\begin{tabular}[t]{@{\hspace{0em}}p{\colOneWidth} p{\colTwoWidth}}
            \texttt{BoltHoles} \\
            \hspace*{1em}\texttt{ListOf(BoltHole)}  
        \end{tabular}

        &
         A list of bolt holes which should be included in the generated profile.
        \\
        \\


        \begin{tabular}[t]{@{\hspace{0em}}p{\colOneWidth} p{\colTwoWidth}}
            \texttt{EndOfPrompt} 
        \end{tabular}

        &
        A special token denoting the end of the geometric prompt.
        \\
        
        \bottomrule
    \end{tabular}
\end{table}

\subsection{Construction steps}
The construction steps are introduced in Section \ref{subsection:learned sequences} and a small number of examples given.  The remaining construction steps are listed in Table \ref{tab:construction_steps_contd}.  Each construction step contains an special token describing the step type, then a series of input and output geometry, encoded as described  in Tables \ref{tab:single_token_entities} and \ref{tab:multi_token_entities}.

The list of construction steps starts with a special token \texttt{StartOfConstruction} and ends with token \texttt{EndOfConstruction}.  Prior to construction steps which make use of parameter values, a \texttt{UseParameterN} parameter index token is provided in the sequence as described in Appendix \ref{appendix:parameters}.   As soon as sufficient geometry has been created to produce a curve in the final profile, the a \texttt{CreatedCurve} token is added, followed by the \texttt{BoundLine}, \texttt{Arc} or \texttt{Circle} tokens as described in Table \ref{tab:multi_token_entities}.  We found the introduction of these tokens to be critical to performance of the network.  Without them the profiles predicted at the end of the sequence often contained self-intersection.  The complete construction sequences for a number of generations are shown in full in Appendix \ref{appendix:generated_constrction_sequence_examples}.

\begin{table}[!htb]
    \centering
    \caption{Examples of construction steps continued}
    \label{tab:construction_steps_contd}
    \renewcommand{\arraystretch}{0.5}
    \begin{tabular}{@{}p{5cm} p{5cm} p{3cm}@{}}
        \toprule
        \textbf{Description} & \textbf{Explanation} & \textbf{Example} \\
        \midrule
        \begin{tabular}[t]{@{}l@{}}
        \texttt{CircleReverseCircle} \\
        \textbf{Input:} \texttt{circle\textsubscript{1}} \\
        \textbf{Output:} \texttt{circle\textsubscript{2}}
        \end{tabular}
        &
        Given an oriented circle, return the same circle with the opposite orientation. 
        &
        \raisebox{-0.75\height}{\includegraphics[width=1.9cm]{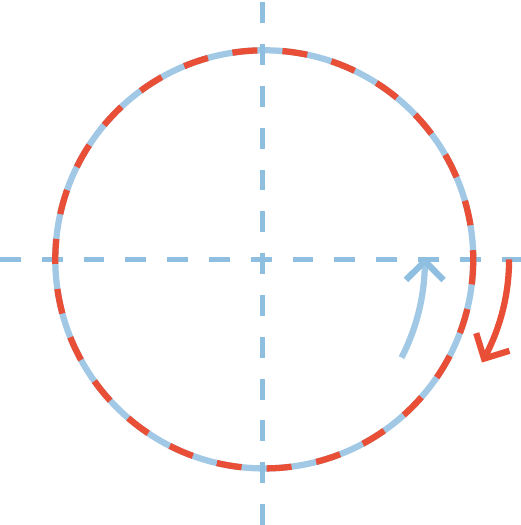}}
        \\
        \midrule
        \begin{tabular}[t]{@{}l@{}}
        \texttt{CirclePointPointArc} \\
        \textbf{Input:} \texttt{circle, point\textsubscript{start},}\\ \texttt{point\textsubscript{end}} \\
        \textbf{Output:} \texttt{point\textsubscript{arc}}
        \end{tabular}
        &
        Given a directed circle and a start and end point on the
        circle.  Find the mid point of the arc which would trim
        the circle to the span which starts at the start point and
        follows the circles direction to the end point.
        &
        \raisebox{-0.75\height}{\includegraphics[width=2cm]{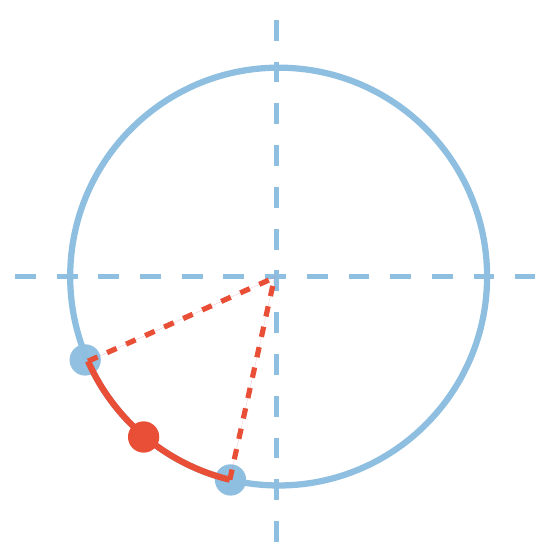}}
        \\
        \midrule
        \begin{tabular}[t]{@{}l@{}}
        \texttt{LineDatumParallelLine} \\
        \textbf{Input:} \texttt{line\textsubscript{1}, point\textsubscript{datum},} \\
        \textbf{Output:} \texttt{line\textsubscript{2}}
        \end{tabular}
        &
        Given a line and a datum point, find and return the line running
        parallel to the given line passing through the datum
        point.
        &
        \raisebox{-0.75\height}{\includegraphics[width=2.2cm]{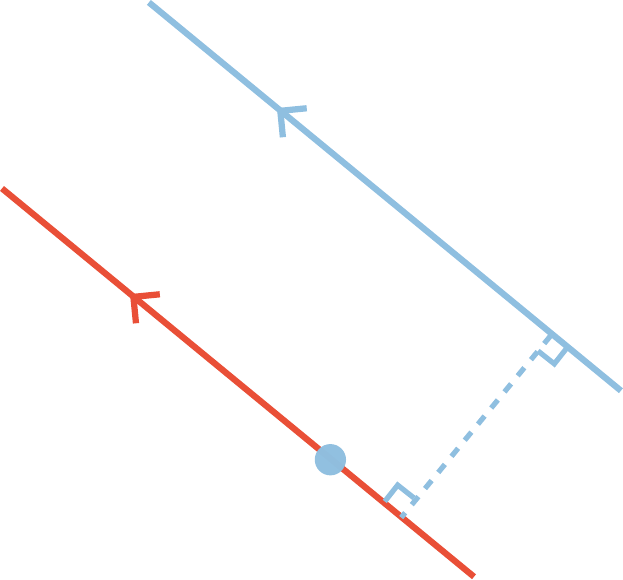}}
        \\
        \midrule
        \begin{tabular}[t]{@{}l@{}}
        \texttt{LineLineFillet} \\
        \textbf{Input:} \texttt{line\textsubscript{1}, line\textsubscript{2},} \\
                        \texttt{radius}\\
        \textbf{Output:} \texttt{arc}
        \end{tabular}
        &
        Given two directed lines and a radius, find and return the fillet
        arc which can be created along with its ordered start and
        end point.
        &
        \raisebox{-0.75\height}{\includegraphics[width=2cm]{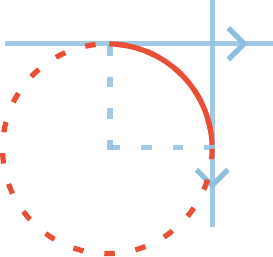}}
        \\
        \midrule
        \begin{tabular}[t]{@{}l@{}}
        \texttt{LineCircleParallelLine} \\
        \textbf{Input:} \texttt{line\textsubscript{1}, circle\textsubscript{1}} \\
        \textbf{Output:} \texttt{line\textsubscript{2}}
        \end{tabular}
        &
        Given an oriented line and a circle, compute the line parallel to the original that is tangent to the circle. There are exactly two such lines, one on each side of the circle, the orientation of the input line determines which one is selected.
        &
        \raisebox{-0.75\height}{\includegraphics[width=2cm]{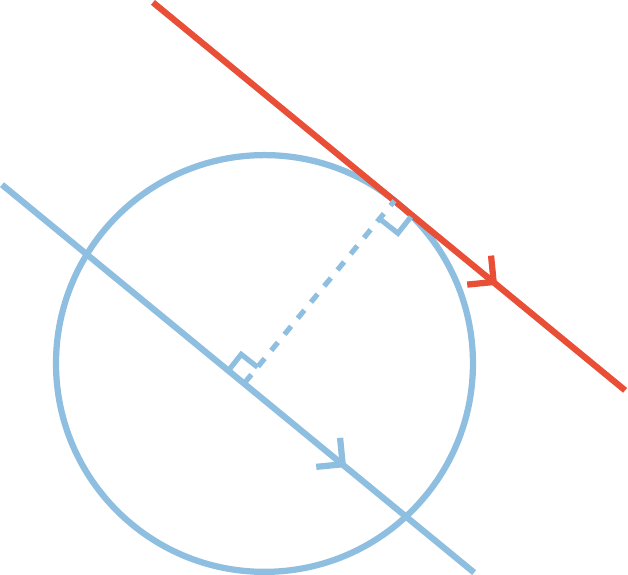}}
        \\
        \midrule
        \begin{tabular}[t]{@{}l@{}}
        \texttt{LineSymLineLine} \\
        \textbf{Input:} \texttt{line\textsubscript{1}, line\textsubscript{sym}} \\
        \textbf{Output:} \texttt{line\textsubscript{2}}
        \end{tabular}
        &
        Given a line and a symmetry line, find and return the image of the line reflected across the symmetry line.
        &
        \raisebox{-0.75\height}{\includegraphics[width=2cm]{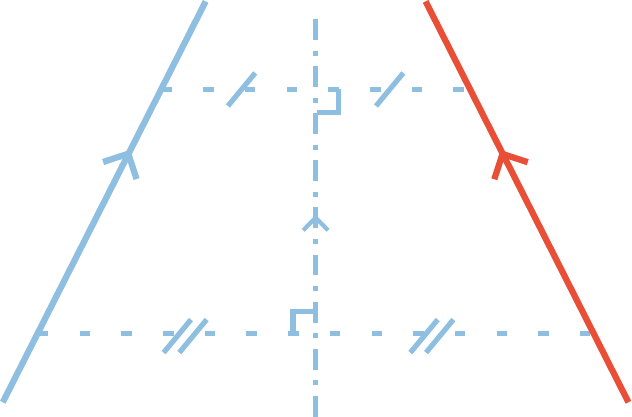}}
        \\

        \midrule
        \begin{tabular}[t]{@{}l@{}}
        \texttt{PointLineSymPointStep} \\
        \textbf{Input:} \texttt{point\textsubscript{1}, line\textsubscript{sym}} \\
        \textbf{Output:} \texttt{point\textsubscript{2}}
        \end{tabular}
        &
        Given a point and a symmetry line, find and return the image of the point reflected across the symmetry line.
        &
        \raisebox{-0.75\height}{\includegraphics[width=2cm]{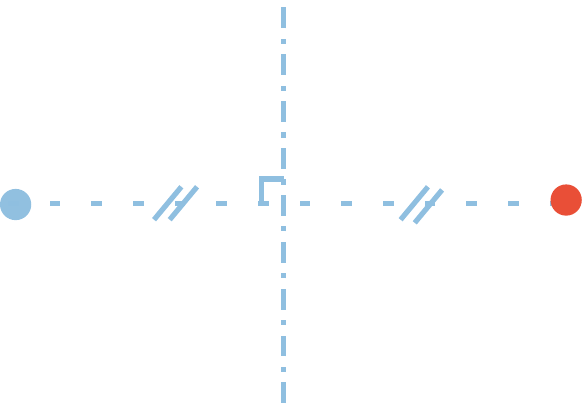}}
        \\

        \midrule
        \begin{tabular}[t]{@{}l@{}}
        \texttt{LineReverseLine} \\
        \textbf{Input:} \texttt{line\textsubscript{1}} \\
        \textbf{Output:} \texttt{line\textsubscript{2}}
        \end{tabular}
        &
        Given a directed line, return it with reversed direction.
        &
        \raisebox{-0.7\height}{\includegraphics[width=2cm]{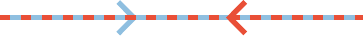}}
        \\
        \midrule
        \begin{tabular}[t]{@{}l@{}}
        \texttt{LineAxisRotatedLine} \\
        \textbf{Input:} \texttt{line\textsubscript{1}, angle} \\
        \textbf{Output:} \texttt{line\textsubscript{2}}
        \end{tabular}
        &
        Given a line a point and an angle, construct and return
        a line rotated by an angle in either the clockwise or
        counter-clockwise direction.
        &
        \raisebox{-0.75\height}{\includegraphics[width=1.9cm]{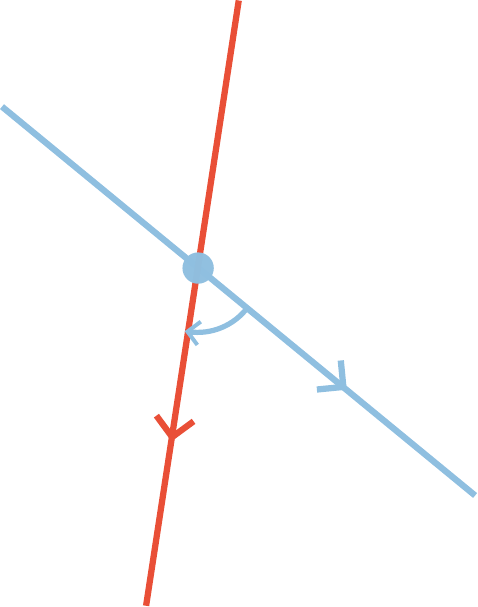}}
        \\
        \midrule
        \begin{tabular}[t]{@{}l@{}}
        \texttt{PointRadiusCircle} \\
        \textbf{Input:} \texttt{point, radius} \\
        \textbf{Output:} \texttt{circle}
        \end{tabular}
        &
        Given a point and a radius, create and return a circle.
        &
        \raisebox{-0.75\height}{\includegraphics[width=2cm]{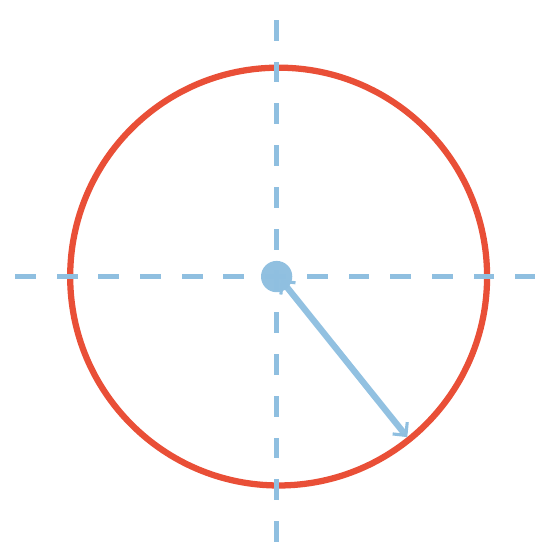}}
        \\

        \bottomrule
    \end{tabular}
\end{table}


\subsection{Parameters}
\label{appendix:parameters}
The construction sequences reduce the degrees of freedom in the generated profile to a small number of parameters which can be used for interactive shape editing.  The values of these parameters are predicted by the model, but can be adjusted by the designer and the sequences replayed with floating point precision.  Before any parameter value is used in a construction step, a special \texttt{UseParameterN} tag token is included indicating the index of the parameter to be used. There are $32$ separate  tokens, \texttt{UseParameter0}, \texttt{UseParameter1}, ...,  allowing up to 32 seaparate parameters to be defined.  After this token the parameter value is given.   This will either be a \texttt{Length} or \texttt{Angle} token indicating the parameter value.

\subsection{Profile geometry}
\label{appendix:profile_geometry}

The sequences finish with the profile geometry, which is encoded using a domain specific language similar to \cite{wu2021deepcad, xu2022skexgen}.  In loops consisting of multiple arc and line segments, the start point of each curve is the end point of the previous curve.  Lines can then be defined by their end points and arcs represented by mid points and end points.  A special token indicates the start of a new loop.  As circles are closed curves, they are always in separate loops and are represented by a center point and radius.  The tokenization for the profile geometry is shown in Table \ref{tab:profile_geometry}.

\begin{table}[!ht]
    \centering
    \caption{The tokenization of the final profile.  As profiles contain only closed loops, the end point of the previous curve can be used as the start point of the current cure}
    \label{tab:profile_geometry}
    \renewcommand{\arraystretch}{0.5}
    \begin{tabular}{@{}p{0.35\textwidth} p{0.65\textwidth}}
        \toprule
        \textbf{Tokens} & \textbf{Explanation}  \\

        \midrule
        \begin{tabular}[t]{@{\hspace{0em}}p{0.3\linewidth} p{0.6\linewidth}}
            \texttt{StartOfProfile}
        \end{tabular}

        &
        A special token indicating the start of the profile
        \\
        
        \midrule
        \begin{tabular}[t]{@{\hspace{0em}}p{0.3\linewidth} p{0.6\linewidth}}
            \texttt{PolyArcLineLoop}
        \end{tabular}

        &
        Indicates the start of a loop containing multiple lines and arcs
        \\

        \midrule
        \begin{tabular}[t]{@{\hspace{0em}}p{0.3\linewidth} p{0.6\linewidth}}
          \texttt{SingleCircleLoop} & \\
          \hspace*{1em}\texttt{Point}  & // \emph{Center point} \\
          \hspace*{1em}\texttt{Length} & // \emph{Radius}
        \end{tabular}

        &
        Indicates a loop comprising a single circle
        \\
        
        \midrule
        \begin{tabular}[t]{@{\hspace{0em}}p{0.3\linewidth} p{0.6\linewidth}}
            \texttt{ProfileLine} \\
            \hspace*{1em}\texttt{Point} &//  \emph{End point}
        \end{tabular}

        &
        A profile line defined by it's end point.  The end point of the previous curve is used as this line's start point.
        \\

        \midrule
        \begin{tabular}[t]{@{\hspace{0em}}p{0.3\linewidth} p{0.6\linewidth}}
          \texttt{ProfileArc} & \\
          \hspace*{1em}\texttt{Point} & // \emph{Mid point} \\
          \hspace*{1em}\texttt{Point} & // \emph{End point}
        \end{tabular}

        &
        A profile arc defined by it's mid point and end point.  The end point of the previous curve is used as this arcs's start point.
        \\
        
        \midrule
        \begin{tabular}[t]{@{\hspace{0em}}p{0.3\linewidth} p{0.6\linewidth}}
            \texttt{EndOfProfile}
        \end{tabular}

        &
        A special token indicating the end  of the profile data
        \\

        \bottomrule
    \end{tabular}
\end{table}

\section{Extracting construction sequences from profile geometry}
\label{appendix:profile2prompt_algorithm}

\subsection{Pre-processing}
Before the extraction of the construction sequences starts, the profiles are normalized so they fit into a unit square centered on the origin.  This allows a consistent set of tolerances to be used in the downstream processing.  A length tolerance is defined as $1/127$ model units, matching the size of the quantization bins used for tokenization.  Points are considered coincident if the distance between them is smaller than this tolerance.   Vectors are considered parallel when they deviate by less than 1 degree.   

The following pre-processing is then applied.
\begin{itemize}
    \item Curves shorter than the length tolerance are removed.  The end points of adjacent curves are moved to close the gaps.
    \item Collinear line segments are merged
    \item The curves in each loop are lexicographically sorted, based on the end point of line and arc segments.
    \item Loops are ordered with the outer loop first and inner loops lexicographically sorted.  For multi-curve loops the sort is conducted using using the end point from the first curve.  For circles, the bottom point is used.
\end{itemize}

\subsection{Extracting geometric prompts} \label{appendix:appendix_geometric_prompts}

Rather than training our model to auto-complete profiles from a random subset of curves as in \cite{Seff2021VitruvionAG}, we select the geometry to include in our geometric prompts based on common design requirements.  The geometric information extracted and the reasoning for extracting it is described below.

\noindent\textbf{Line segments}: When extrusions are combined to build a 3d model, they often align with one another at planar faces.  Consequently when defining profile geometry, it's useful to be able to specify line segments which will become a part of the convex hull of the generated profile.  To construct the geometric prompt, we begin by identifying all line segments on the convex hull. From these, we select a subset, aiming to preserve groups of symmetrical segments. Groups that can be removed without reducing the convex hull of the line set are then discarded at random.


\noindent\textbf{Bolt holes}: Bolts are commonly used to connect mechanical components together.  As such, it is often necessary for mechanical components to contain bolt holes at specific locations.  We identify bolt holes in profiles as circular inner loops.   These are specified by the hole center, a radius representing the size of the hole and a clearance disk which defines the required strength of the part around the bolt.   Bolts can fail by tearing out of the bolt hole if the amount of material around them is insufficient to withstand transverse forces, hence the size of the clearance around bolts is an important factor designers want to control. 

\noindent\textbf{Profile area}:  This allows the thickness of the generated profile to be controlled, allowing the physical strength of the part to be balanced against weight and material utilization. 

\noindent\textbf{Fraction of tangent continuous vertices}:  Allowing designers to control the proportion of tangent-continuous vertices lets them decide whether the generated shape will have sharp or rounded corners. Tangent continuous vertices can be easily detected by comparing the tangents at the ends of adjacent curves.
    
\noindent\textbf{Symmetry lines}:  Many mechanical parts exhibit symmetries and the ability to enforce symmetry is required by designers.  Symmetries are identified using a 2d implementation of \cite{Li2015Symmetry} and incorporated into the geometric prompt as infinite lines. 

\noindent\textbf{Bounding box}:   Parts will be generated in a unit square and then scaled and positioned when imported into the CAD application.  An intuitive way for designers to specify the size and aspect ratio for the generated profile is using an oriented bounding box in the coordinate system of the CAD software.  This can be mapped to a 2d bounding box which fits snugly into the unit square.  The generated profile can then be transformed and scaled back into the CAD systems coordinates.   The generative model can also exploit the aspect ratio of the bounding box as an additional shape control.  

\noindent\textbf{Center of gravity}:  In cases where the desired profile covers only a small fraction of the bounding box area, and is poorly defined by other geometry in the prompt, the center of gravity can be used to guide the shape.  This control is especially useful when the shape needs to avoid regions which would lead to clashes in the assembly context.

\noindent\textbf{Profile complexity}:  It's often desirable to have control over the complexity of the profile.  Generative algorithms can create unwanted  additional detail when the complexity to too high, and can give very simple shapes when the complexity is too low.  The number of edges in the  profile is used as a complexity measure in the geometric prompt.

\noindent\textbf{Number of internal loops}:   Designers often want to create designs which are both strong and lightweight.  By providing control over the number of internal loops in the profile, in conjunction with the profile area, we can encourage the model to produce additional internal loops to lightweight the shape.

\noindent\textbf{Dimensioning datum}: Designers often want to parameterize profiles based on some fixed geometry.  In parametric CAD this is often the fixed origin of the coordinate system or a reference point created by a previous modeling feature.  To allow designers control over this fixed geometry, we include a datum point in the geometric prompt.   The datum point used is randomly selected from the following:
\begin{itemize}
    \item A corner of the profile's bounding box
    \item A mid point of one of the bounding box edges
    \item The center of the bounding box
    \item The center of the largest circle in the profile
\end{itemize}
Some examples of how the geometric prompt controls the shape can be seen in Appendix \ref{appendix:shape_control_from_geometric_prompt}

\subsection{Construction sequence extraction}

An overview of the construction sequence extraction algorithm was given in Section \ref{subsection:dataset_creation}.   Here we give more detail on each phase of the algorithm.

\subsubsection{Analysis phase}
The profile analysis phase is similar to the first step of a heuristic auto-constrainer.  The geometry of the profile curves are analyzed and the following groups of geometric entities are detected
\begin{itemize}
    \item Groups of collinear points
    \item Groups of parallel lines
    \item Fillet arcs and their adjacent curves
    \item Symmetry lines and the points and curves placed symmetrically about them  
    \item Groups of concentric circles or arcs
\end{itemize}

\subsubsection{Identification of frequently used distances}
\label{appendix:frequently_used_distances}
The following distances between pairs curves are recorded
\begin{itemize}
    \item The distance between all pairs of parallel lines
    \item The difference in radius between all pairs of concentric circles
    \item The radii of all circles and arcs
\end{itemize}

The frequency with which these values are found is then counted.   Length parameters are created preferentially for length values which appear more frequently.

\subsubsection{Identification of source lines}
The construction sequences are designed to start with geometry defined in the prompt and, step by step, construct the geometry needed to build the profile.   To enable this we identify the following ``source lines", from which subsequent construction lines will be derived by a series of offsets.   
\begin{itemize}
    \item Symmetry lines.
    \item Infinite lines derived from all line segments provided in the prompt.
    \item A horizontal and a vertical line passing through the dimension datum.
    \item Any group of parallel lines which is not parallel or anti-parallel to one of the source lines identified so far must then be accounted for.  A \texttt{LineAxisRotatedLine} step is created which rotates the closest coordinate system axis line around the dimension datum to produce a new source line parallel to the lines in the group. 
\end{itemize}

\subsubsection{Identification of offsets between parallel lines}
\label{appendix:offsets_between_parallel_lines}
Given a set of directed parallel lines, it is possible to construct any line from any other given the appropriate signed offset distance.   Groups of parallel lines are processed individually to discover chains of offsetting operations which best construct the group.

Pairs of lines which can be constructed from a symmetric offset of a symmetry line are identified and these symmetric constructions are used preferentially.  

The algorithm then tries to describe every line in the remaining group by a sequence of offsets starting from the “source lines”.  An undirected fully connected graph is constructed with lines as the nodes and the distances between lines as the edges.  Edges are then deleted from this graph, starting with the edges corresponding to the least frequent distances identified above.   Edges cannot be removed if doing so would make the graph disjoint.  Finally, the order in which lines are created is defined.  We start with the source lines and traverse the  graph, following the Dijkstra shortest path algorithm.  This traversal defines which lines are the input and output of \texttt{LineOffsetLine} steps. 

\subsubsection{Identification of offsets between circles}
 Similar to the algorithm for parallel lines described above, offsetting operations between concentric circles/arcs are created based on the most frequent length values in the entire profile.    Each group of concentric circles includes at least one operation which creates the initial circle from a center point and radius.
 
\subsubsection{Creating the construction steps}
Next an extensive list of construction steps, which encompasses all the relationships identified between geometric entities is created.  The criteria for creating the steps is as follows

\noindent\textbf{\texttt{LineXLine}}:
In places where a vertex of the profile is coincident with the intersection of two lines, a \texttt{LineXLine} step is created.  The directions of the input lines always follows the directions of the corresponding line segments in the profile.

\noindent\textbf{\texttt{LineLineFilletStep}}:
In places where a fillet arc exists between two line segments, a \texttt{LineLineFillet} step is created

\noindent\textbf{\texttt{PointLineSymPoint}}:
In places where a vertex in the profile can be defined by mirroring another vertex across a symmetry line, a pair \texttt{PointLineSymPoint} step are created.  The decision as to which point is the original and which point will be mirrored is made later in the graph simplification phase.  At this stage two operations are created using the two different points as the input.

\noindent\textbf{\texttt{LineOffsetLine}}:
Line offsetting steps can be created for each line and offset found by the procedure described in Appendix \ref{appendix:offsets_between_parallel_lines}

\noindent\textbf{\texttt{LineDatumParallelLineStep}}:
In cases where a line is require, which is parallel to a ``source line" and passes thorough one of the end points of a  line segment given in the prompt, a \texttt{LineDatumParallelLineStep} is created.   As these steps do not introduce new parameter values they will be used in preference to \texttt{LineOffsetLine} steps.

\noindent\textbf{\texttt{LineCircleParallelLine}}: 
A list of circles is built which includes the clearance disks of all prompt holes along with any  circles or arcs in the profile.   Lines in the profile which touch these circles tangentially are then found.  Steps are created which offset a source line to a position which touches the circle.   There are two solutions for tangentially touching lines and the direction of the initial line is used to select between them.  As above, these steps do not introduce new parameter values, causing them to be selected preferentially.

\noindent\textbf{\texttt{CirclePointPointArc}}:
For each arc in the profile, which is not defined by a fillet, a step is created which finds the arc’s mid-point from the underlying circle and the start and end points of the arc.  A direction flag indicates whether the clockwise or counter clockwise arc is selected.  The introduction of these steps allow arcs to be derived from circles provided the end points are known.

\noindent\textbf{\texttt{LineXCicle}}:
For each profile vertex which is at a non-tangential intersection between a line and an arc, a \texttt{LineXCicle} step is created.

\subsubsection{Graph completion}
The construction steps are then used to build a bipartite dataflow graph with geometry and construction steps as nodes and edges representing the flow of geometry into and out of the operations.  The directed graph is then searched for geometry nodes without incoming edges. Geometry introduced by the prompt is expected not to have incoming edges, while other geometry without incoming edges is problematic.  New construction steps need to be added to the graph to allow this geometry to be constructed from the prompt.    In many cases lines or circles can be created from existing geometry by introducing \texttt{LineReverseLine} or \texttt{CircleReverseCircle} steps.  Circles can also be created using \texttt{PointRadiusCircle} steps provided the center point has been constructed.   Graphs which contain geometry without incoming edges following the graph pruning phase will be discarded. 

 \subsubsection{Parameters}
The \texttt{LineOffsetLine}, \texttt{CircleOffsetCircle}, \texttt{PointRadiusCircle}, \texttt{LineLineFilletStep} steps require input length parameters and the \texttt{LineAxisRotatedLine} steps requires an input angle parameter.  The required parameters are identified and special geometry nodes are created to represent their values.  A single parameter can be an argument to multiple construction steps, for example, it's common for multiple fillets to have the same radius.

\subsubsection{Cycle breaking}

The initial dataflow graph will contain cycles, which must be broken before processing continues.  Cycles in the graph are detected by a depth first search algorithm and processed one at a time.  The list of edges forming each cycle is found and the edge to remove is chosen based on the following criteria:

\noindent\textbf{Essential edges}: Starting from the geometry required to define the final profile we examine the incoming edges.  If only one edge is incoming then this geometry is derived from the output of a single construction step.   The edge must then be added to an “essential edges” set.  We then move to the construction step which gave rise to these each of the essential edges and repeat this procedure.   The entire graph is traversed and the complete set of essential edges found.  The traversal stops when we reach a geometry node with more than one incoming edge.   We cannot remove any essential edges from the graph.

\noindent\textbf{Hops from prompt}: We want to find the shortest possible construction sequences for each piece of geometry.  Starting at the prompt geometry we use  Dijkstra's algorithm to find the smallest number of hops  to each node in the graph.  As the graph contains cycle, care must be taken to avoid visiting nodes more than once.  

\noindent\textbf{Construction step priorities}:
Construction steps have the following priorities.
\begin{itemize}
    \item Creating arcs as fillets is highest priority
    \item Next creating lines which are tangent to circles
    \item Next creating points from line circle intersections
    \item Then operations mirroring lines or points
    \item Finally all other operations
\end{itemize}

The cycles are then broken by removing one edge from each cycle.  Essential edges are never removed.  The non-essential edges which have nodes with the largest number of hops from prompt geometry are removed in preference.   The priorities of the operations at the tail end of each edge is then used as a tie breaker.

\subsubsection{Graph pruning}

Geometry nodes with more than one incoming edge are equivalent to over-constrained but consistent geometry in a traditional constraint solver.   The graph pruning algorithm selects one incoming edge for each geometry and removes branches of the graph which lead to other solutions.
The choice is made based on a cost function which is evaluated for each edge in the graph.   
\begin{itemize}
    \item Geometry provided in the prompt has a cost of zero
    \item Construction steps add a cost of 0.1 to all output edges
    \item Parameters incur a cost of 1.0.
    \item Any geometry which is not defined by the prompt and has no predecessor nodes is given an infinite cost
\end{itemize}
	
Once the costs for the edges have been computed, we loop over the geometry nodes.  For geometry node with more than one incoming edge, the edge with the smallest cost is kept and all others will be removed.   Branches of the graph which do not end at profile geometry are then removed.  Profiles are discarded from the dataset if any geometry required to build the final shape could not be constructed from the geometry in the prompt.

\subsubsection{Sequence ordering}

The dataflow graph now contains all the information required to create the construction sequence.  The nodes in the graph are first ordered using a lexicographical topological sort.  This respects the dependencies between geometry and construction steps defined by the graph while allowing ambiguity in the ordering to be resolved by secondary criteria.   The order in which curves appear in the final profile is the primary criteria used and a lexicographical ordering of the intermediate geometry is used to break ties.  Infinite lines are ordered based on the lexicographically lowest point where they intersect the bounding box of the profile.

The parameters required to define the shape are then ordered based on the first construction step which utilizes them.  Each parameter is allocated an index based on this order, which is used in the  \texttt{UseParameterN} tokens described in Appendix \ref{appendix:parameters}. 

\section{Additional results}
\label{appendix::additinal_results}

\subsection{RL algorithms}

\begin{figure}[H]
    \centering
    \includegraphics[width=0.5\textwidth]{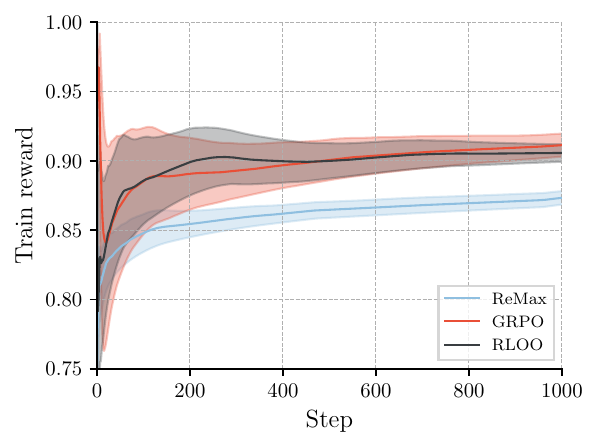}
    \caption{RL training rewards}
    \label{fig:rl_training}
\end{figure}

The hyper-parameters used for RL model alignment are detailed in Table~\ref{tab:RL_hparams}. Training was performed on 3 NVIDIA A10 GPUs with the AdamW optimizer and a learning rate of $1\times10^{-6}$ for one thousand steps. Figure \ref{fig:rl_training} shows the training performance over steps for the different reinforcement learning algorithms used. Applying a KL penalty proved essential. In its absence, the model exploited the reward function by consistently generating invalid sequences. Since rewards for invalid sequences were masked, only generations for extremely simple prompts yielded nonzero rewards, leading to a trivially high average reward despite no meaningful learning.

\begin{table}[!htb]
    
    \centering
    \caption{Hyperparameter comparison across different RL algorithms.}
    \label{tab:RL_hparams}
    \begin{tabular}{@{} lccc @{}}
        \toprule
        \textbf{Hyperparameters} & \textbf{ReMax} & \textbf{GRPO} & \textbf{RLOO} \\
        \midrule
        Effective batch size     & 768           & 144           & 144          \\
        \midrule
        Group sample size & -- & 16 & 16 \\ \midrule
        Learning rate & 1e-6 & 1e-6 & 1e-6 \\ \midrule
        Top-p sampling & 0.3 & 0.3 & 0.3 \\ \midrule
        Token-level KL penalty & 0.1 & -- & 0.1 \\ \midrule
        Sequence-level KL penalty & -- & 0.01 & -- \\ \midrule
        Policy clipping ratio & -- & 0.2 & -- \\
        \bottomrule
    \end{tabular}
\end{table}

\subsection{Construction step precision}
As each construction step represents an atomic operation with well defined output geometry, we can measure the accuracy with which the transformer learns to perform the construction steps.  These values are shown for all construction step models in Table \ref{tab:constrction_step_metrics}.  Distances are measured in the unit square in which the profile is constructed.  Angles are measured in radians and comparisons between boolean of flags are 1.0 when the flags agree and 0.0 when they disagree.  All values are averaged over the 6427 prompts in the test set.

We observe that the transformer can learn to exactly duplicate tokens with perfect accuracy.  For example, when a circle is reversed, only the \texttt{IsCCW} flag changes and  perfect agreement is observed for the center point and radius.  The precision for construction steps with a single token as input and output is also very good, hence the good agreement in distance from the origin and angle when lines are reversed.

More complex geometric operations, with multiple curves are arguments have bigger errors.  Examples of this are line-line and line-circle intersections (\texttt{LineXLine} and \texttt{LineXCircle\textsubscript{dist}})  and the creation of the mid point of fillet arcs (\texttt{LineLineFillet\textsubscript{precision}}).

Applying the reinforcement learning algorithms to the sequences has only a small effect.   This implies the network primarily learns how to perform the geometric operations in supervised pre-training. The RL rewards then help the network to combine the individual operations into CAD programs which give rise to syntactically and geometrically valid profiles rather than improving its ability to perform the low level geometric computations.

\begin{table}[!htb]
    \centering
    \caption{Comparison of construction step metrics among different models}
    \label{tab:constrction_step_metrics}
    \begin{tabular}{@{} l c c c c @{}}
        \toprule
        \textbf{Metrics} &
        \makecell{\textbf{Construction} \\ \textbf{steps model}} &
        \makecell{\textbf{Construction} \\ \textbf{steps model}\\ \textbf{(ReMax)}} &
        \makecell{\textbf{Construction} \\ \textbf{steps model}\\ \textbf{(GRPO)}} &
        \makecell{\textbf{Construction} \\ \textbf{steps model}\\ \textbf{(RLOO)}} \\
        \midrule
        \texttt{CircleOffsetCircle\textsubscript{center}} ($\downarrow$) & \textbf{0.0000} & \textbf{0.0000} & \textbf{0.0000} & \textbf{0.0000} \\
        \midrule
        \texttt{CircleOffsetCircle\textsubscript{radius}} ($\downarrow$) & 0.0109 & \textbf{0.0107} & 0.0124 & 0.0133 \\
        \midrule
        \texttt{CirclePointPointArc\textsubscript{dist}} ($\downarrow$) & 0.1413 & 0.1438 & 0.1413 & \textbf{0.1102} \\
        \midrule
        \texttt{CircleReverseCircle\textsubscript{ccw}} ($\uparrow$) & \textbf{1.0000} & \textbf{1.0000} & \textbf{1.0000} & \textbf{1.0000} \\
        \midrule
        \texttt{CircleReverseCircle\textsubscript{center}} ($\downarrow$)  & \textbf{0.0000} & \textbf{0.0000} & \textbf{0.0000} & \textbf{0.0000} \\
        \midrule
        \texttt{CircleReverseCircle\textsubscript{radius}} ($\downarrow$)  & \textbf{0.0000} & \textbf{0.0000} & \textbf{0.0000} & \textbf{0.0000} \\
        \midrule
        \texttt{SolutionExists} ($\uparrow$) & 0.9981 & 0.9983 & 0.9987 & \textbf{0.9988} \\
        \midrule
        \texttt{LineAxisRotatedLine\textsubscript{angle}} ($\downarrow$) & \textbf{0.0204} & 0.0242 & 0.0243 & 0.0226 \\
        \midrule
        \texttt{LineAxisRotatedLine\textsubscript{dist}} ($\downarrow$) & \textbf{0.0130} & 0.0148 & 0.0132 & 0.0139 \\
        \midrule
        \texttt{LineCircleParallelLine\textsubscript{angle}} ($\downarrow$) & 0.0017 & 0.0059 & 0.0020 & \textbf{0.0009} \\
        \midrule
        \texttt{LineCircleParallelLine\textsubscript{dist}} ($\downarrow$) & 0.0239 & 0.0248 & 0.0236 & \textbf{0.0231} \\
        \midrule
        \texttt{LineDatumParallelLine\textsubscript{angle}} ($\downarrow$) & 0.0020 & 0.0019 & 0.0022 & \textbf{0.0017} \\
        \midrule
        \texttt{LineDatumParallelLine\textsubscript{dist}} ($\downarrow$) & \textbf{0.0061} & 0.0063 & 0.0066 & 0.0066 \\
        \midrule
        \texttt{LineLineFillet\textsubscript{precision}} ($\downarrow$) & \textbf{0.0347} & 0.0367 & 0.0362 & 0.0365 \\
        \midrule
        \texttt{LineLineFillet\textsubscript{valid\_circle}} ($\uparrow$) & 0.9320 & 0.9372 & \textbf{0.9522} & 0.9504 \\
        \midrule
        \texttt{LineOffsetLine\textsubscript{angle}} ($\downarrow$) & 0.0040 & 0.0040 & \textbf{0.0039} & \textbf{0.0039} \\
        \midrule
        \texttt{LineOffsetLine\textsubscript{dist}} ($\downarrow$) & 0.0095 & 0.0097 & 0.0095 & \textbf{0.0094} \\
        \midrule
        \texttt{LineReverseLine\textsubscript{angle}} ($\downarrow$) & 0.0015 & 0.0007 & \textbf{0.0006} & 0.0007 \\
        \midrule
        \texttt{LineReverseLine\textsubscript{dist}} ($\downarrow$) & \textbf{0.0001} & \textbf{0.0001} & \textbf{0.0001} & \textbf{0.0001} \\
        \midrule
        \texttt{LineXCircle\textsubscript{dist}} ($\downarrow$) & 0.0866 & 0.0870 & \textbf{0.0745} & 0.0787 \\
        \midrule
        \texttt{LineXLine} ($\downarrow$) & 0.0245 & 0.0203 & 0.0194 & \textbf{0.0184} \\
        \midrule
        \texttt{PointLineSymPoint} ($\downarrow$) & 0.0060 & 0.0066 & 0.0057 & \textbf{0.0049} \\
        \midrule
        \texttt{PointRadiusCircle\textsubscript{center}} ($\downarrow$)  & \textbf{0.0000} & \textbf{0.0000} & \textbf{0.0000} & \textbf{0.0000} \\
        \midrule
        \texttt{PointRadiusCircle\textsubscript{radius}} ($\downarrow$) & \textbf{0.0000} & \textbf{0.0000} & \textbf{0.0000} & \textbf{0.0000} \\
        \midrule
        \texttt{SymlineOffsetLineLine\textsubscript{angle}} ($\downarrow$) & \textbf{0.0003} & 0.0004 & \textbf{0.0003} & 0.0005 \\
        \midrule
        \texttt{SymlineOffsetLineLine\textsubscript{dist}} ($\downarrow$) & \textbf{0.0055} & \textbf{0.0055} & 0.0056 & 0.0056 \\
        \bottomrule
    \end{tabular}
\end{table}
\section{Generated construction sequence examples}
\label{appendix:generated_constrction_sequence_examples}
In this section we show some additional construction sequences generated by the model and relate these to the parametric behavior of the shape when the driving parameters are varied.  

\begin{figure}[H]
    \centering
    \includegraphics[width=\textwidth]{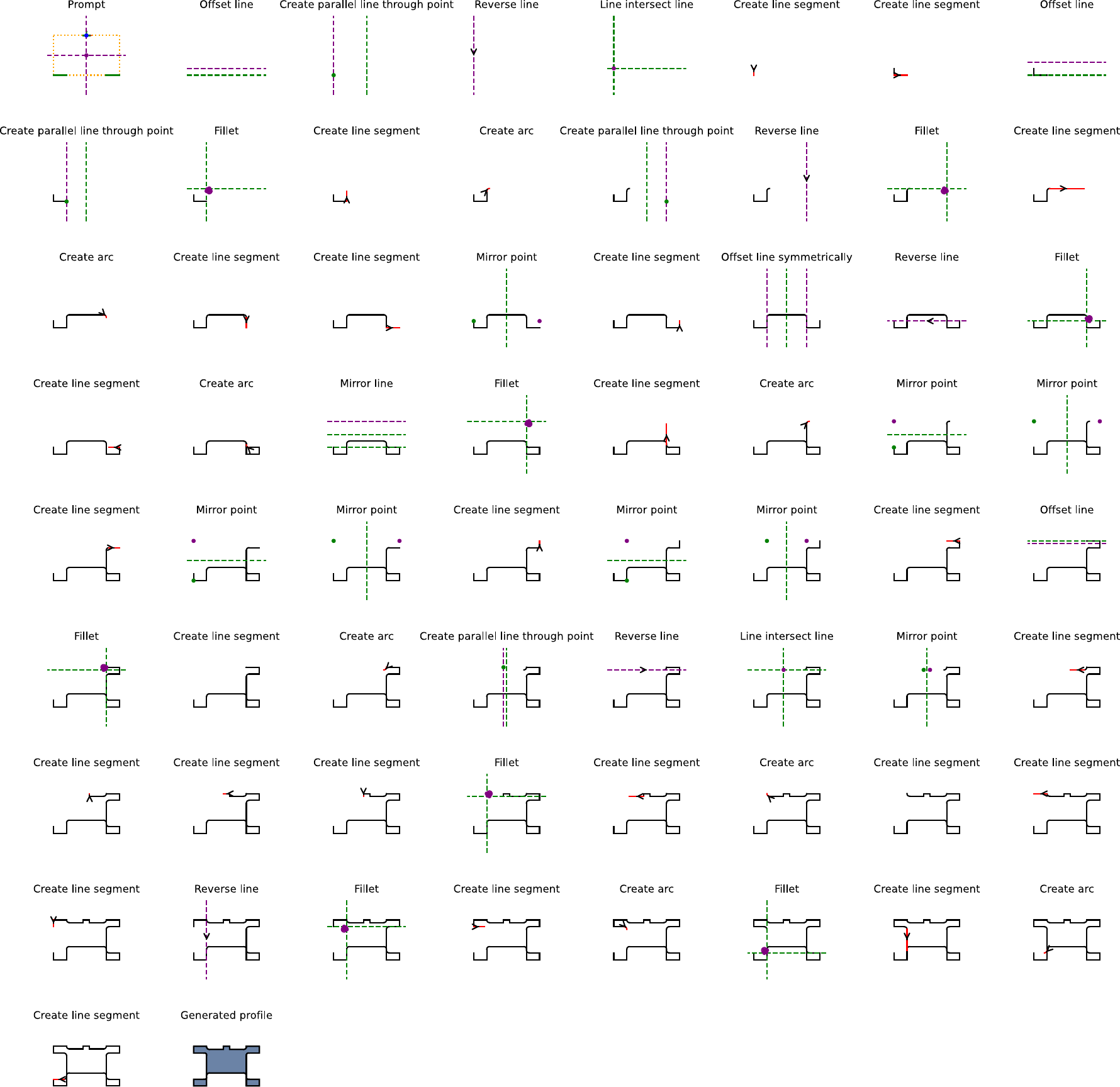}
    \caption{Construction sequence for the example with both horizontal and vertical lines.  Points present in the prompt are mirrored over the two symmetry lines and used to build the final shape.}
      \label{fig:geom_prompt_symmetry_construction_sequence}
\end{figure}

\begin{figure}[H]
    \centering
    \includegraphics[width=\textwidth]{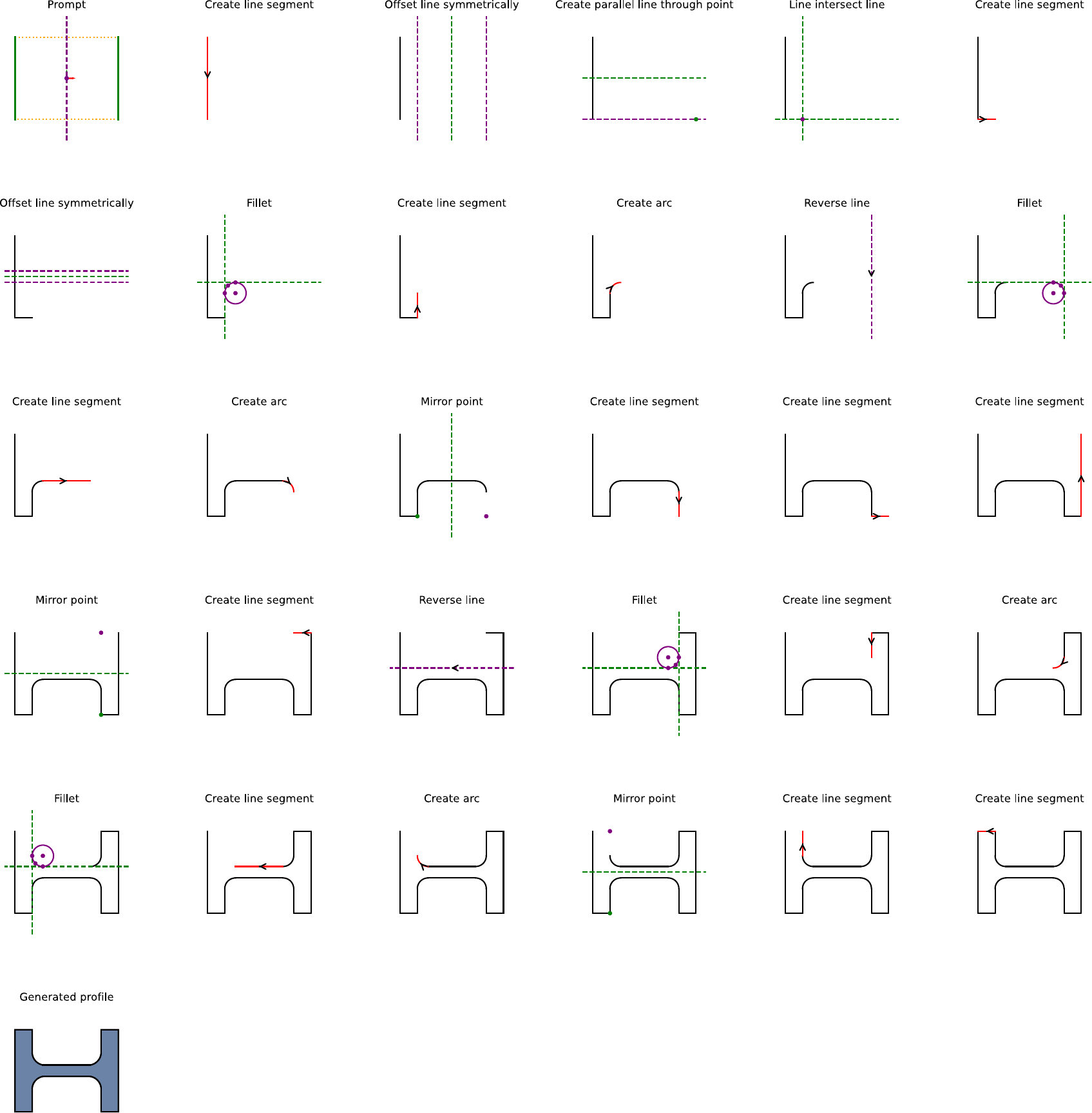}
    \caption{Construction sequence for an I-beam shape. Notice how the vertical symmetry line is defined, but the central position of the dimension datum causes a horizontal symmetry line to be implied.}
\end{figure}

\begin{figure}[H]
    \centering
    \includegraphics[width=\textwidth]{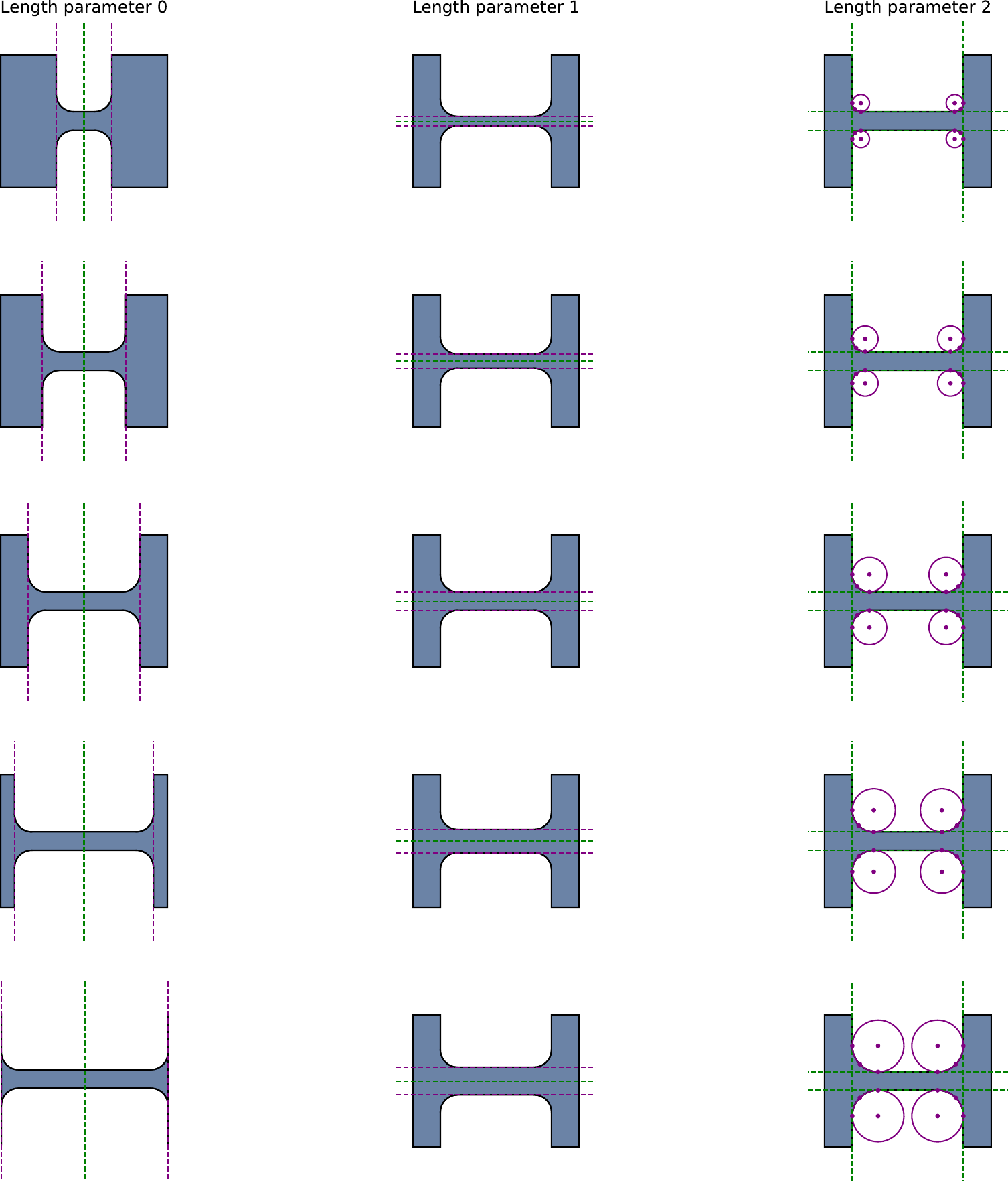}
    \caption{Parametric variations for the I-beam. The profile is parameterized with three values which allow control over the width of the two side bars, the width of the central beam and the radii of the four fillet arcs.}
\end{figure}

\begin{figure}[H]
    \centering
    \includegraphics[width=\textwidth]{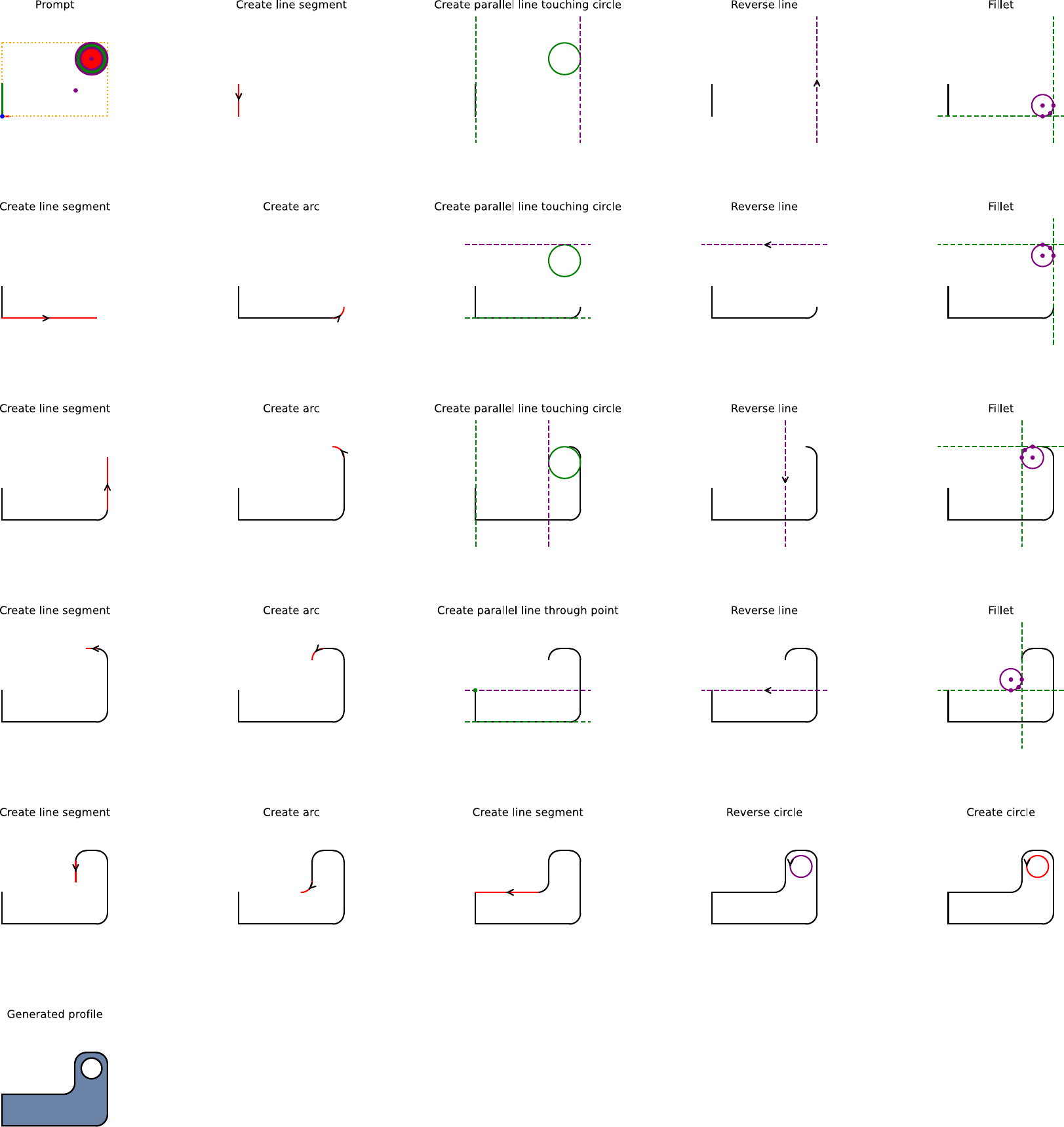}
    \caption{An example of a construction sequence with no free parameters.  Notice how the vertical lines are created by offsetting the axis lines until they touch the clearance disk.  The  horizontal lines are derived from the end points of the line segment provided in the prompt and the dimension datum.}
\end{figure}

\begin{figure}[H]
    \centering
    \includegraphics[width=\textwidth]{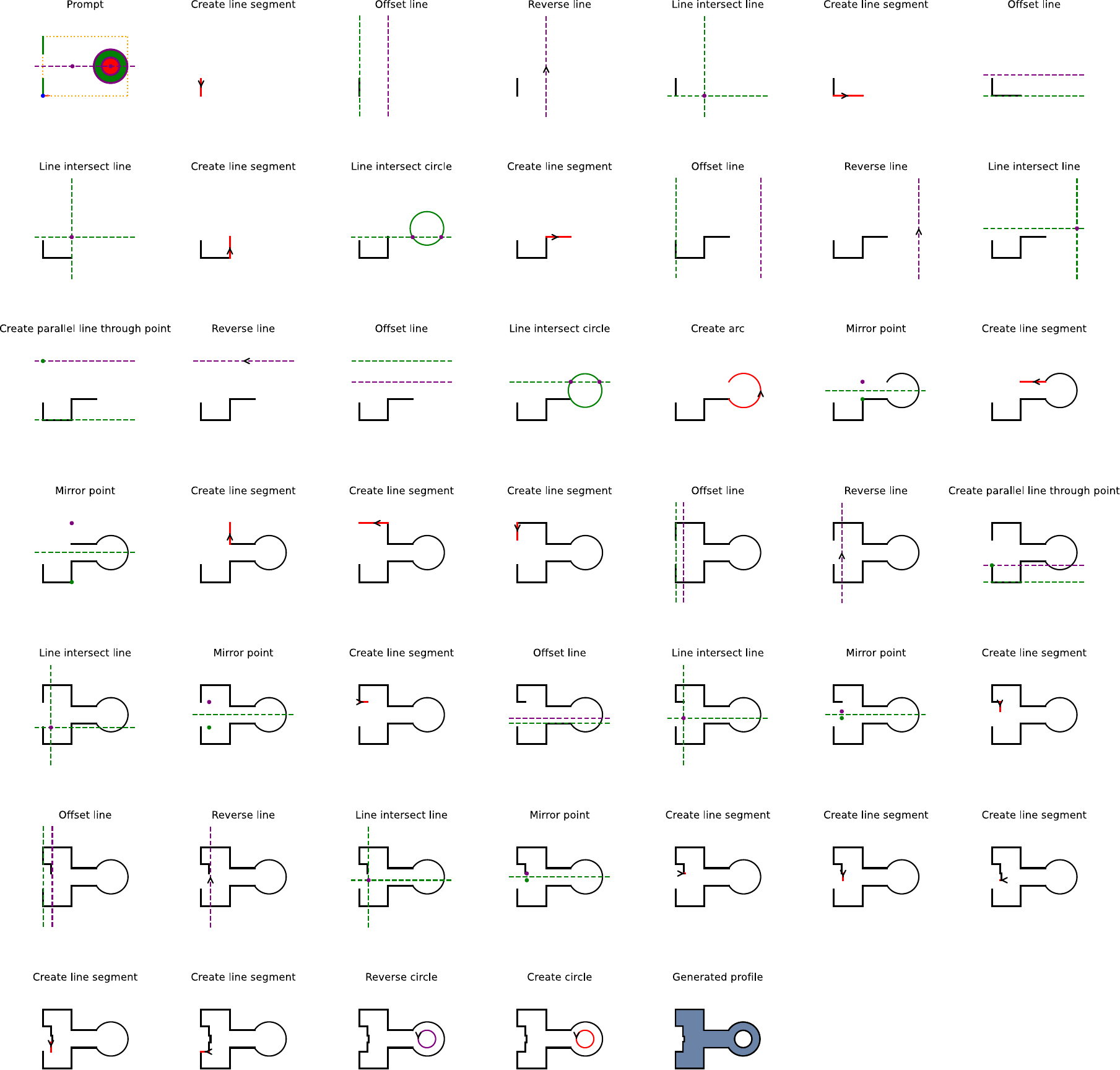}
    \caption{A construction sequence for a more complex shape.}
\end{figure}

\begin{figure}[H]
    \centering
    \includegraphics[width=\textwidth]{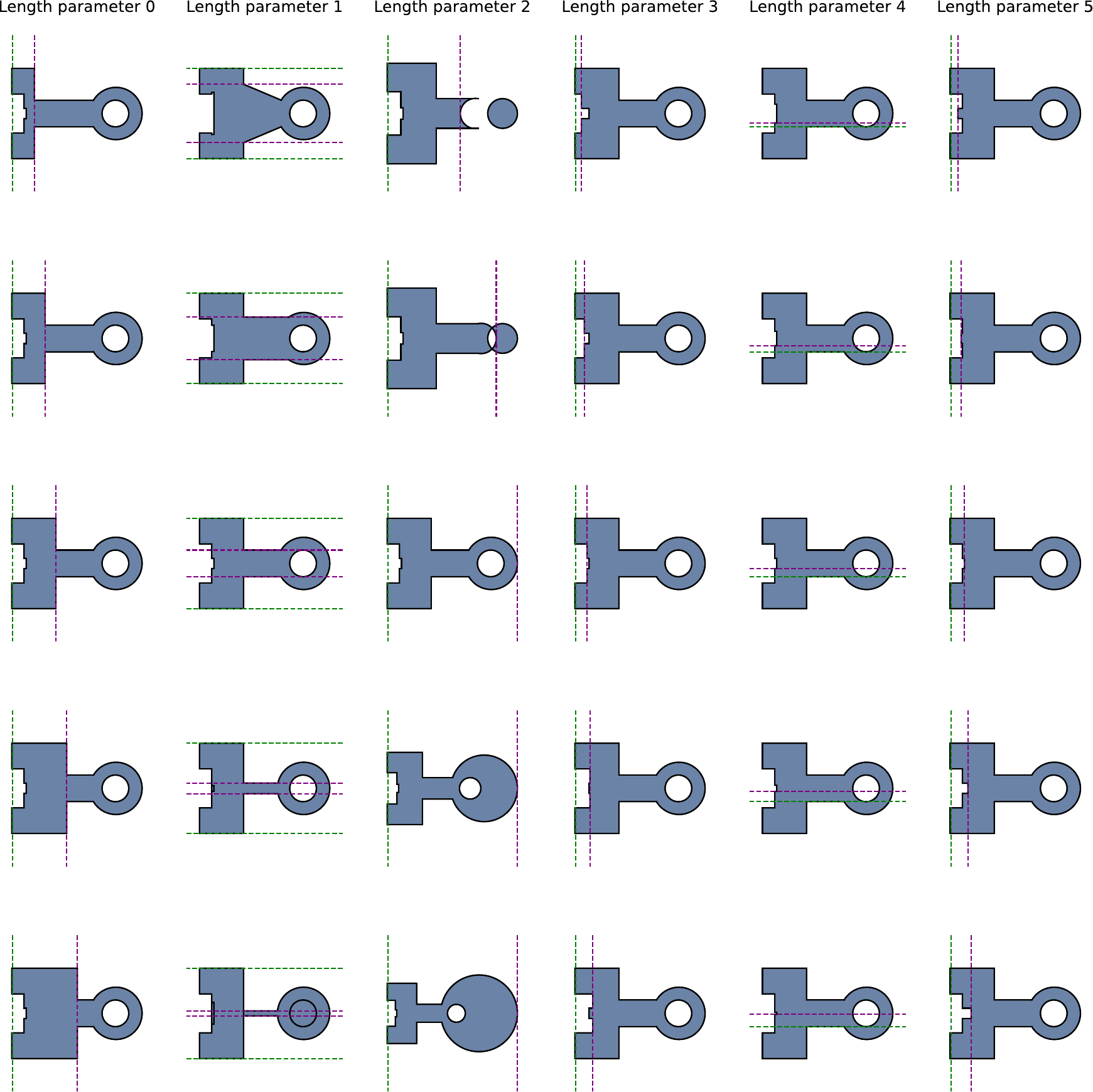}
    \caption{Parametric variations for the complex shape.  We see that the parameterization is sensible in most cases, however this example does show two failure modes.  When parameter 1 causes the line segments to move to a position where they fail to intersect with the circle then this operation fails.  The points are not updated causing the introduction of sloped lines.  Parameter 2, defines a line offset which is intended to be at the right hand side of the shape.  This line is intersected with the center line to build the mid point of the outer circular arc.  When this line is moved, the outer arc and inner circle are no longer concentric, revealing a problem with the parameterization.  Techniques which improve the parametric behavior of the resulting profiles are left to further work.}
\end{figure}

\end{document}